\documentclass[journal]{new-aiaa}
\usepackage[utf8]{inputenc}
\let\checkmark\undefined
\usepackage{dingbat}
\usepackage{bbding}
\usepackage{amssymb}  
\usepackage{graphicx}
\usepackage{amsmath}
\usepackage{wrapfig}
\usepackage{lipsum}
\usepackage[version=4]{mhchem}
\usepackage{siunitx}
\usepackage{longtable,tabularx}
\usepackage[table]{xcolor}
\usepackage{rotating}

\title{Overview of Guidance, Navigation and Control System of the TeamIndus lunar lander}

\author{Vishesh Vatsal$^*$, C. Barath$^*$, J. Yogeshwaran$^*$, Deepana Gandhi$^*$, Chhavilata Sahu$^*$, Karthic Balasubramanian$^*$, Shyam Mohan$^*$, Midhun S. Menon \footnote{Guidance, Navigation and Control Engineer, TeamIndus, Axiom Research Labs, Survey No. 9, Off Bellary Road, Jakkuru Main Rd, Jakkuru Layout, Byatarayanapura, Bengaluru, Karnataka, India 560092}, P. Natarajan \footnote{Guidance, Navigation and Control Senior Expert, TeamIndus, Axiom Research Labs, Survey No. 9, Off Bellary Road, Jakkuru Main Rd, Jakkuru Layout, Byatarayanapura, Bengaluru, Karnataka, India 560092}, Vivek Raghavan \footnote{Technical Manager, TeamIndus, Axiom Research Labs, Survey No. 9, Off Bellary Road, Jakkuru Main Rd, Jakkuru Layout, Byatarayanapura, Bengaluru, Karnataka, India 560092}} 
\affil{Guidance, Navigation and Control team, Axiom Research Labs, Survey No. 9, Off Bellary Road, Jakkuru Main Rd, Jakkuru Layout, Byatarayanapura, Bengaluru, Karnataka, India 560092}

\begin{document}

\maketitle

\begin{abstract}
TeamIndus’ Lunar Logistics vision includes multiple lunar missions to meet requirements of science, commercial and efforts towards global exploration. The first mission is slated for launch in 2020. The prime objective is to demonstrate autonomous precision lunar landing, and Surface Exploration Rover to collect data on the vicinity of the landing site. TeamIndus has developed various technologies towards lowering the access barrier to the lunar surface. This paper shall provide an overview of design of lander GNC system.

The design of the GNC system has been described after concluding studies on sensor and actuator configurations. Frugal design approach is followed in the selection of GNC hardware. The paper describes the constraints for the orbital maneuvers and lunar descent strategy. Various aspects of the GNC design of autonomous lunar descent maneuver - time line of events, guidance, inertial and optical terrain relative navigation schemes are described. The GNC software description focuses on system architecture, modes of operation, and core elements of the GNC software. The GNC algorithms have been validated using Monte-Carlo simulations and processor-in-loop testing. The paper concludes with a summary of key risk mitigation studies for soft landing.

\end{abstract}

\newpage

\section{Introduction}
\lettrine{T}{he} genesis of the TeamIndus (TI) lunar lander mission was the Google Lunar X Prize (GLXP) which called on privately funded teams to land on the Moon, travel 500 meters and send back high definition imagery. Guidance, Navigation and Control (GNC) is a key enabling system for the mission and is responsible for assuring Sun pointing and burn execution during the Orbital segment and delivering the lander to a desired landing site on the Moon. Several recent papers have given an overview of GNC design for precision robotic lunar landers \cite{lee2010preliminary}\cite{li2016guidance} \cite{fisackerly2011esa}. To increase the probability of success, some of the key features that landers have incorporated in new designs are Vision Based Absolute Navigation \cite{singh2008lunar}, Hazard Avoidance Algorithms in Terminal Descent \cite{johnson2008analysis} and Terrain Relative Navigation \cite{johnson2008overview} through an advanced sensor suite. However, there is not sufficient literature on low-budget lunar lander missions with minimum redundancy. The key requirements, trade studies that helped to evolve the current GNC system architecture are described. GNC mode definitions for orbits and descent highlights the degree of autonomy selected for the mission. Modeling and simulation results help gain confidence in the mission. The paper concludes with some studies concerned with reducing risks to mission success during lunar descent.

\section{The mission profile}
 
\begin{wrapfigure}{r}{0.6\textwidth}
  \centering
   \vspace{-5mm}
  \includegraphics[trim=10mm 8mm 1mm 30mm, clip,width=0.6\textwidth]{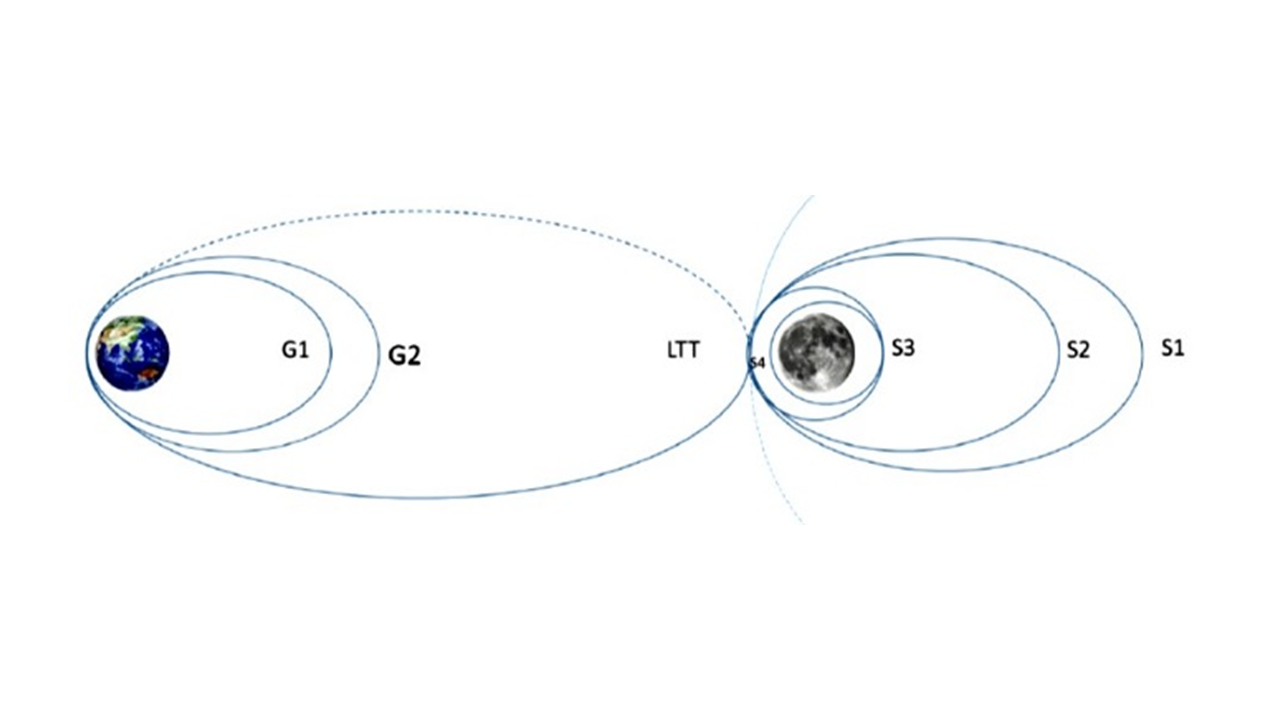}
  \vspace{-20mm}
  \caption{Mission Profile.}
  \vspace{-5mm}
\end{wrapfigure}

\par The vehicle is injected into an initial Earth orbit (G1) by the launch vehicle. The first significant burn, the Translunar Injection (TLI) maneuver targets a perilune altitude of 100 km at lunar capture and puts the spacecraft on a minimum energy lunar transfer trajectory (LTT) to the Moon. Trajectory Correction Maneuvers (TCMs) are planned on the first and third day after TLI to ensure arrival conditions at orbit capture. The Lunar Orbit Capture (LOC) maneuver delivers the vehicle to an orbit of 100 x 6700 km (S1) selected on the basis of a stability analysis. Since the mission targets an early morning landing, the landing time after lunar capture is dictated by the location of the dawn terminator w.r.t. the landing site. This time is spent in a parking orbit (S2). The Parking orbit insertion (POI) burn followed by Circularization burn (CRC) burn results in a nearly circular 100 x 100 km orbit stable orbit (S3). Descent Orbit Insertion (DOI) burn is initiated at the apolune to meet Pre-Descent entry conditions and results in a 12.6 x 100 km orbit (S4). Following a pre-descent sequence, lunar descent is initiated at the third perilune pass. The autonomous powered lunar descent maneuver reduces the vehicle’s orbital velocity of approximately 1.7 km/s to near zero terrain relative speeds. When very close to the lunar surface, the engines are cut off and the vehicle falls onto the lunar surface under lunar gravity. 

\section{Critical GNC requirements}

The GNC system requirements were derived from the higher level GLXP problem statement. These requirements provide specifications for attitude knowledge, pointing errors, redundancy and landing accuracy. Some key requirements driving the GNC design are:
\begin{itemize}
\item	The GNC system shall be able to control attitude at all points of the mission: 15 deg error in Sun pointing and 0.2 deg during any significant propulsive maneuver.
\item	GNC sensor selection shall be done to have redundancy only when the device does not have space heritage. This minimizes the risk of sensor failure under the selection strategy with limited commercial and mass budgets.
\item	The GNC system shall be able to determine terrain relative lateral velocity knowledge errower with <= 0.1 m/s error.
\item	The GNC system shall be able to identify, in real-time, a safe landing location during Terminal Descent.
\item	The GNC system shall be capable of delivering the lander to the desired landing site with a maximum dispersion of 1 km.
\item	The GNC system shall include the capability for always ensuring power positive condition of the spacecraft during the Orbital segment of the mission.
\end{itemize}
\section{Trade studies}

The GNC system design is heavily driven by lunar descent requirements. Hence, system studies were conducted to evolve the structure and parameters of the descent strategy and the hardware configuration required for realization. 

\subsection{Orbit Selection for Descent Initiation}

Following the choice of an orbital descent strategy (no direct descent from LTT), parameters of the descent orbit were studied. It was found \cite{ramanan2005An} that frozen orbits, for which there is no long-term change in eccentricity and argument of periapsis should be having altitudes 100 km or more. Low lunar orbits (orbits below 100 km) suffer from gravitational perturbation effects that make orbits unstable. The perilune altitude was selected with the following constraints: (a) gravity anomalies should not disturb the path of the spacecraft significantly (b) the long-range laser rangefinder with a maximum ranging ability of 16 km must be able to read a slant range at perilune and (c) the orbit path should not intersect with the mountainous terrain. Thus, a 12.6 km altitude or a periapsis of 1750 km was chosen as the point for initiating lunar descent.

\subsection{Actuator Configuration}

The orbital burns and descent segment account for the majority of propellant consumption. A tight mass budget required the mission profile to be fuel-optimal and this drove the actuator configuration. Reaction Wheels (RW) were rejected early on based on mass and power considerations, in favour of thruster-based configurations. It was found that any mass benefit due to the use of RW during the orbital segment were found to be lost during descent since RW acted as dead weight. Solid motor-based configurations were rejected due to low Isp and unavailability of high thrust engines, leaving the final actuator configuration to be based on liquid propulsion.  Furthermore, mono-propellant systems were rejected due to low thrust and Isp. Trade-offs between various bi-propellant actuator configurations were conducted, with the following evaluation metrics: a)Ability to meet variable thrust demands during Braking and Approach phases.
b) Sufficient torque capacity and thruster redundancy for 3-axis attitude control c) Potential mass impact d)Ability to execute descent with a reduced set of thrusters e) Impact on touchdown speeds. 
\begin{table}[h]
\begin{center}
\begin{tabular}{ m{12em} m{3cm} m{3cm} m{3cm}} 
 \hline
 \hline
 \bf Case & \bf 16 thrusters & \bf 14 thrusters & \bf 12 thrusters\\ 
 \hline
 No. of mitigation schemes & 11 & 10 & 10\\ 
 Reliability post thruster failure &12 thruster config reliability&10 thruster config reliability&8 thruster config reliability\\
 Thrust Range &546N to 739N &546N to 709N &546N to 673N\\
 Comments &Ideal & Meets mission requirements at higher touchdown speed & Touchdown speed is high with single thruster failure, thrust range sufficient.\\
 \hline
 \hline
\end{tabular}
\caption{RCT configuration trade-off for the descent}
\vspace{-10mm}
\end{center}
\end{table}
Table 1 shows the result of a trade-off study for RCT configuration for the descent. Based on the trade-off studies, an actuator configuration consisting of a centrally mounted main engine providing 460 N and 16 x 22N thruster was evolved. Redundancy for attitude control was achieved by allocating two blocks of canted 4 x 22N thrusters. Additionally, the main engine was decided to be operated in pulsed-mode during Terminal Descent to meet thrust requirements to meet touchdown speed requirements.

\subsection{Sensor Configuration}
Similar to the actuator configuration, the sensor configuration was dictated by descent requirements. In keeping with the redundancy philosophy, only sensors without space heritage were provided with backup to account for a single device failure. The primary GNC sensors for the mission are the IMU and the SSU. The devices chosen have significant spaceflight heritage, including usage on multiple landing missions. A triad of analog Sun Sensors (CASS) were also chosen to provide independent attitude knowledge during contingency. The sensor configuration for descent was driven by requirements of altitude knowledge error, touchdown speeds and hazard avoidance. LIDAR, Radar altimeters and Laser rangefinders were considered for providing altitude knowledge. Based on cost, schedule and complexity considerations, a pair of military-grade LLRFs was chosen, with the second device serving as a backup. The LLRF, with a range of 16 km and accuracy of ±1 m provides sufficient altitude knowledge for the mission. In order to meet requirements of touchdown speeds and hazard avoidance, a pair of COTS cameras were chosen, with the second device serving as a backup. A cluster of short-range laser range finders, with accuracy of ± 0.2m was chosen to provide high accuracy altitude measurements during Terminal Descent. The small duration of the descent maneuver (approx. 15 min) provided additional justification for the selection of descent sensors with no spaceflight heritage. Finally, touchdown confirmation through telemetry was decided to be done based on persistence of expected IMU and SLRF measurements, removing the need for a dedicated touchdown detection sensor.

\section{GNC Architecture}
Figures 2 and 3 show the top and bottom view respectively of the TeamIndus spacecraft. The IMU, CASS, and SSU are located on upper side of the main deck. The lower part of the main deck (Figure 3) houses all the actuators. The 16 x 22N RCTs are mounted in concentric rings.
 \begin{figure}[h]
  \begin{minipage}[b]{0.5\linewidth}
    \centering
    \includegraphics[trim=5mm 18mm 5mm 10mm, clip,width=1.2\linewidth]{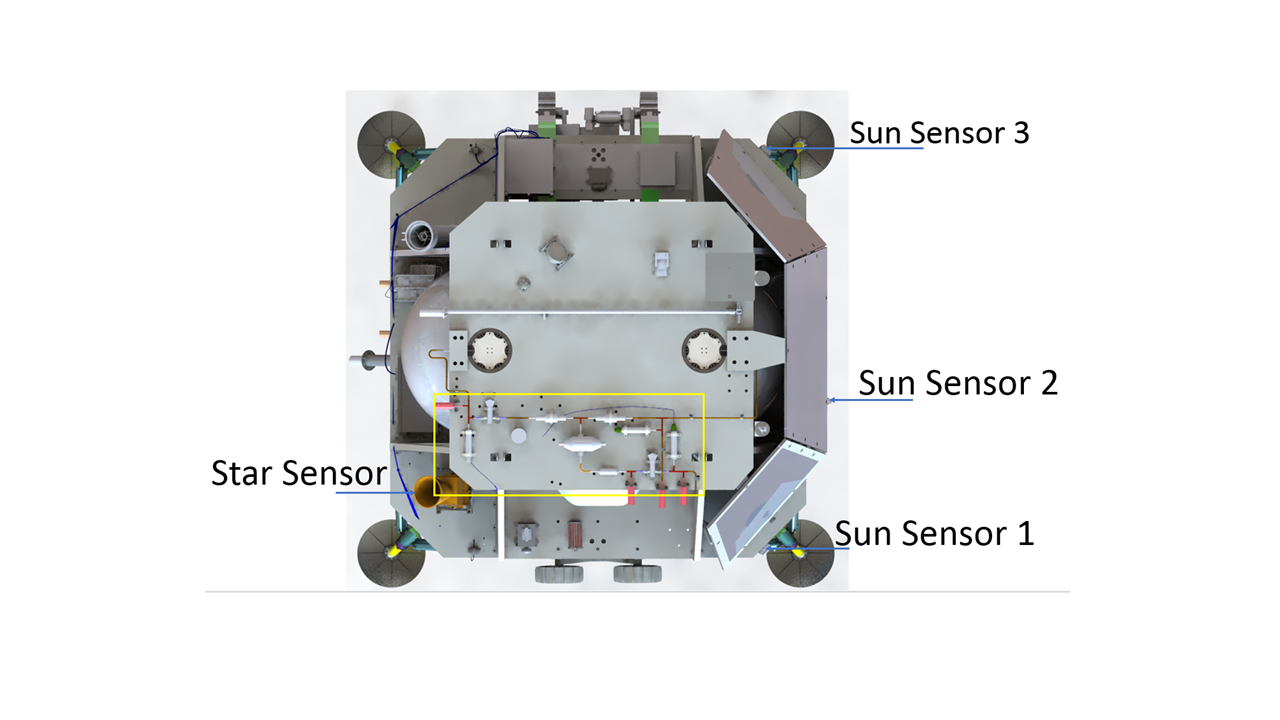}
    \vspace{-8mm}
    \caption{Team
    Indus spacecraft (Top view)}
    
  \end{minipage}
  \hspace{0.5cm}
  \begin{minipage}[b]{0.5\linewidth}
    \centering
    \includegraphics[trim=5mm 18mm 5mm 10mm, clip,width=1.2\linewidth]{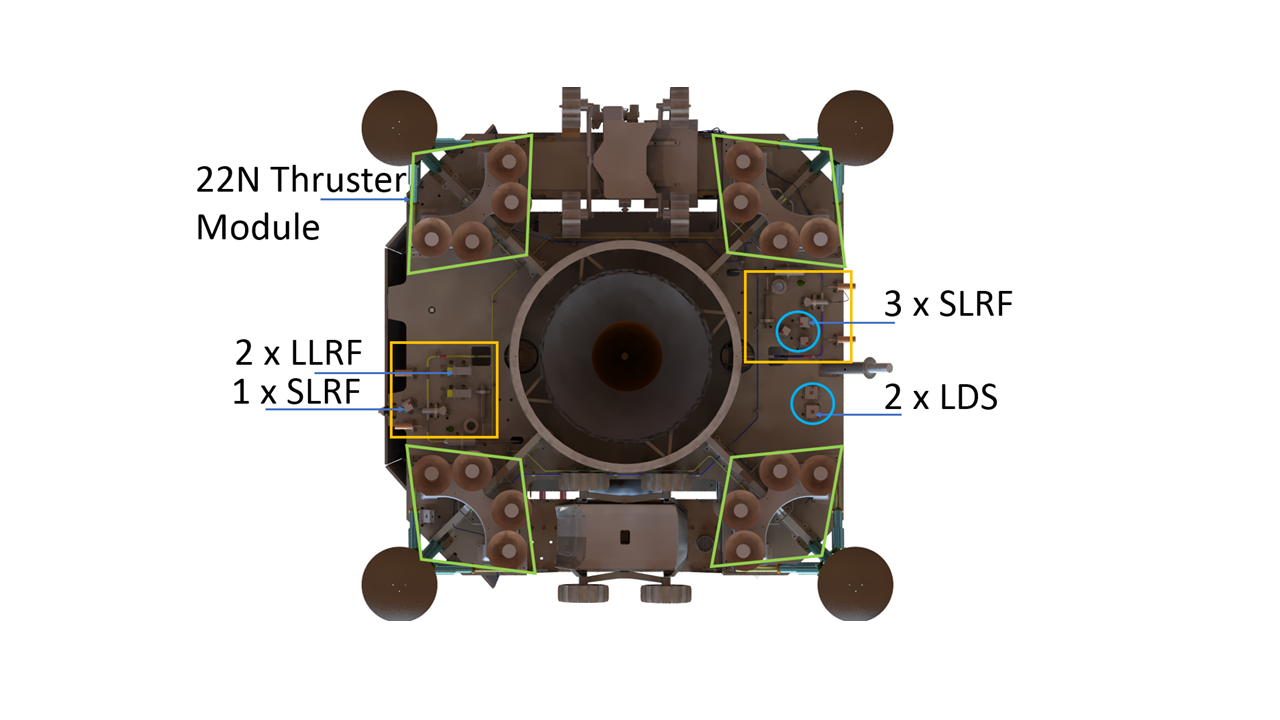}
     \vspace{-8mm}
    \caption{TeamIndus spacecraft (Bottom view)}
   
  \end{minipage}
\end{figure}

 The 8 RCTs in the outer ring have an in-plane canting 10 deg w.r.t. the vehicle longitudinal axis (Z axis) and are grouped into two blocks of 4 RCTs to provide redundant 3-axis attitude control. The other 8 RCTs in the inner ring are aligned to augment the main engine thrust. 

\begin{table}[h]
\begin{center}
\begin{tabular}{ m{18em} m{1cm} m{1cm} m{1cm} m{1cm} m{1cm} m{1cm}} 
 \hline
  \hline
 \bf Function\ Sensors & \bf \begin{turn}{90}1 x Star Sensor\end{turn} & \begin{turn}{90}\bf 3 x Sun sensors\end{turn} & \begin{turn}{90}\bf 1 x IMU\end{turn} &\begin{turn}{90}\bf 2 x LLRF \end{turn}& \begin{turn}{90} \bf 4 x SLRF \end{turn}& \begin{turn}{90} \bf 2 x LDS \end{turn}\\ 
  \hline

 Initial attitude and rate estimation &  \checkmark & & \checkmark  & & & \\
 Initial position and velocity estimation & & & \checkmark  & & & \\

 Propellant mass estimation & & &  \checkmark & & & \\

 The cutoff for deviation from sun pointing direction and safe mode control & & \checkmark  & & & & \\

 Initial state computation for inertial navigation & & & & & & \checkmark  \\

 Touchdown detection & & & \checkmark  & & & \\
 Correction of attitude knowledge & & & &  \checkmark & \checkmark & \\
 Terrain frame definition & & & \checkmark & & \checkmark  & \\

 Terrain- relative lateral velocity estimaton and hazard avoidance & & &  \checkmark & &  \checkmark & \checkmark  \\
 \hline
 \hline

\end{tabular}
\caption{Use cases for GNC sensors}
  \vspace{-15mm}
\end{center}
\end{table}

\begin{wrapfigure}{l}{0.6\textwidth}
  \centering
  \vspace{-5mm}
  \includegraphics[trim=20mm 20mm 0mm 25mm, clip,width=0.7\textwidth]{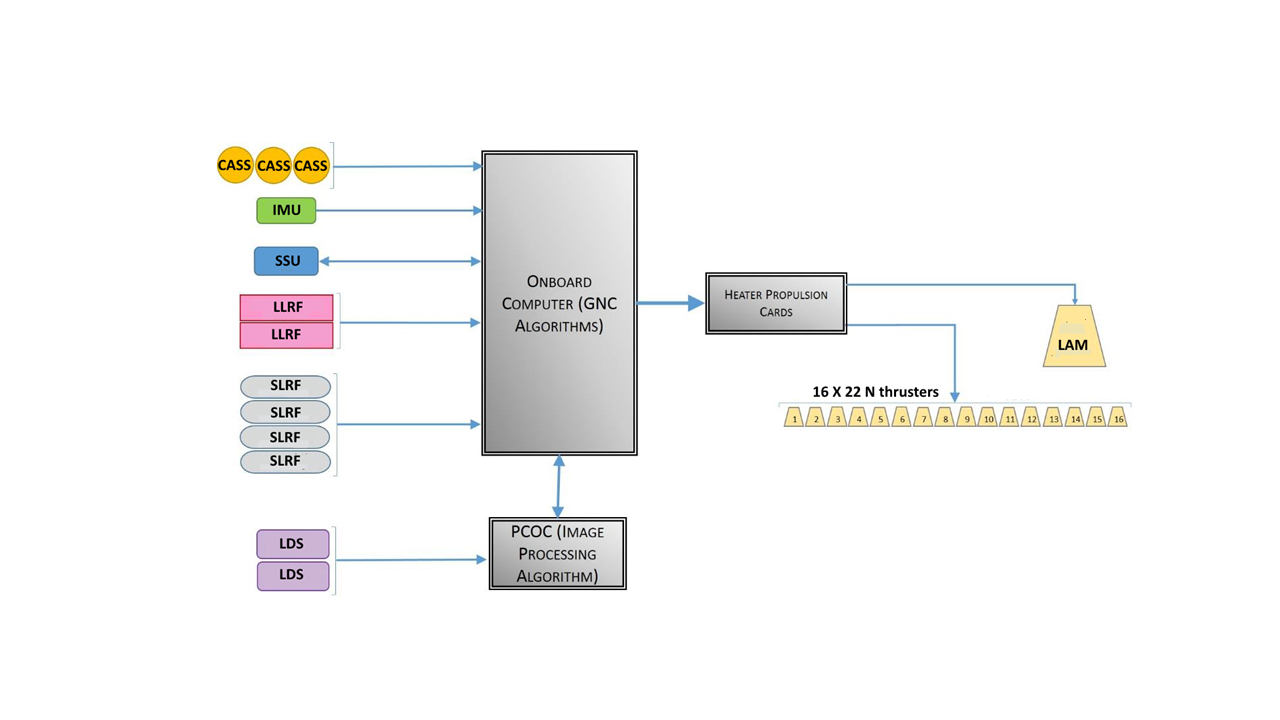}
  \vspace{-8mm}
  \caption{GNC Hardware Interface.}
  \vspace{-5mm}
\end{wrapfigure}

The lower side also houses the 2 x LLRFs, 4 x SLRFs and 2x LDS cameras. The long-range LLRFs are canted 30 deg from the bottom deck to keep the boresight close to the nadir direction during descent.  Table 2 provides a summary of the use cases for the described GNC sensor configuration. Figure 4 shows the GNC hardware interfaces. All sensors except the 2 x LDS are connected to the flight computer (OBC). The OBC selected for the TeamIndus mission utilizes a processor originally designed by ESA and used in multiple missions. All memory devices on the OBC are EDAC enabled. The 2 x LDS are connected to the Payload, Communication and Operations (PCOC) card, which interfaces to the flight computer. The PCOC utilizes a commercially available processor not used in space application before. This too has EDAC capability however implemented in software. This setup allows computationally intensive image processing algorithms to be executed on much faster hardware and also separates time-critical OBC GNC algorithms from dependencies. All actuators are driven through dual Heater Propulsion Cards which interface with the OBC.

\subsection{Software architecture}
\begin{wrapfigure}{l}{0.7\textwidth}
  \centering
  \vspace{5mm}
  \hspace{-15mm}
  \includegraphics[trim=5mm 25mm 0mm 25mm, clip,width=0.8\textwidth]{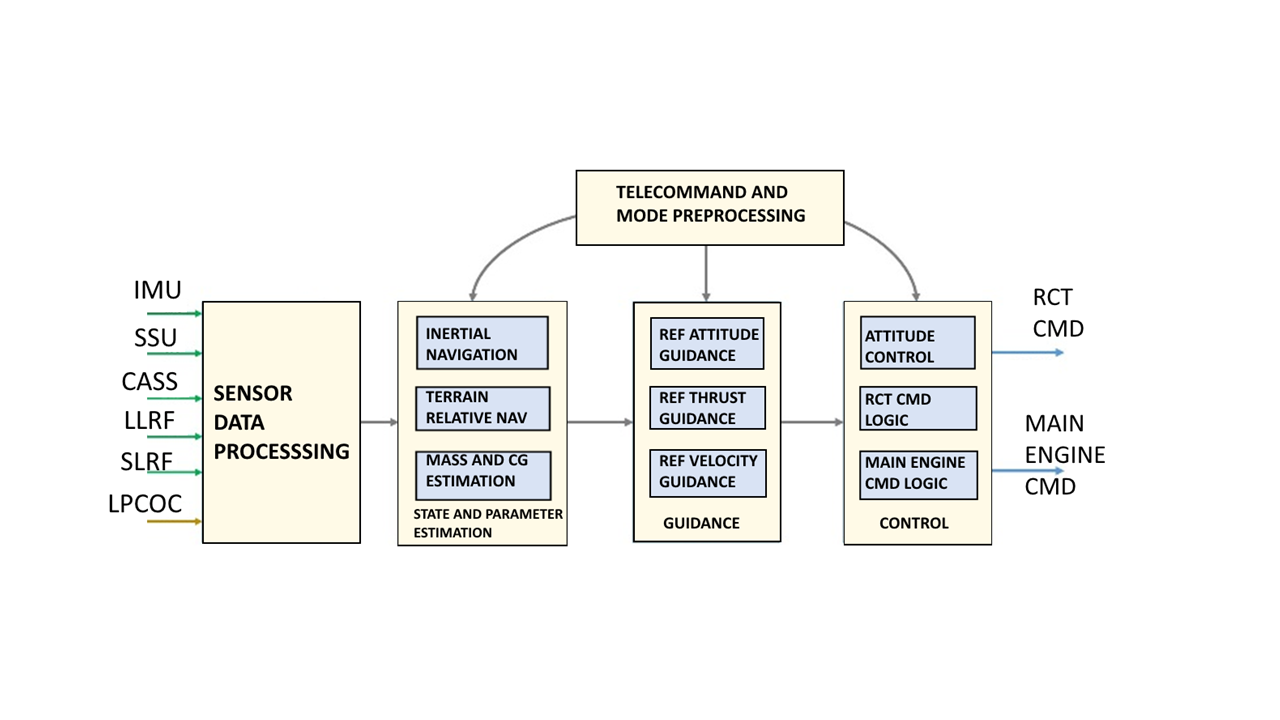}
  \vspace{-15mm}
  \caption{GNC software architecture.}
  \vspace{-5mm}
 \end{wrapfigure}
The GNC software design is modular and provides operational flexibility to effect changes to the GNC software parameters in response to real-time mission events and information, such as during DOI and the pre-descent sequence. As depicted in Figure 5, modules performing GNC functions can be grouped into Sensor Data Processing, State and Parameter Estimation, Guidance and Control. Data processing of all GNC sensors, including from the PCOC is handled by the Sensor Data Processing group. The State and Parameter Estimation group provides attitude knowledge 

 \begin{wrapfigure}{r}{0.6\textwidth}
  \centering
  \vspace{-5mm}
  \hspace{-10mm}
  \includegraphics[trim=65mm 0mm 1mm 1mm,clip,width=1.5\linewidth]{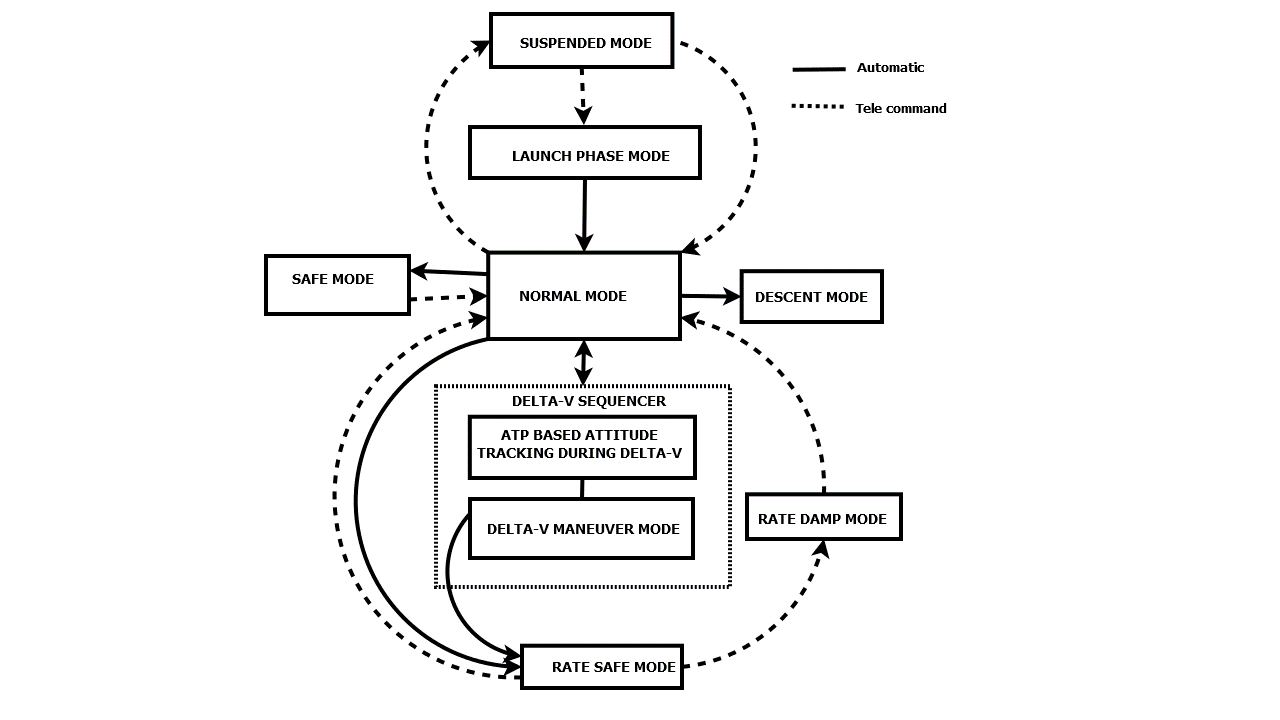}
  \vspace{-10mm}
  \caption{Modes of Operation.}
  \vspace{-5mm}
\end{wrapfigure}

in the orbital segment, and provides attitude, position, velocity and vehicle mass during the descent segment. Attitude Thrust and Velocity command references are provided by modules in the Guidance group. 
 The Control group handles attitude control using a PID controller and PWPFM combination.

 In addition, this group also handles conversion of torque / thrust commands into ON / OFF commands for the 16 RCTs and the Main Engine. The Tele-Command and Mode Processing group handles all Tele-Commands and mode transitions – ground commanded and autonomous and orchestrates the overall software execution. Figure 6 depicts the various modes the lander GNC can operate in. Solid lines indicate autonomous transitions, while dotted lines indicate ground commanded transitions. 
																			The primary orbital mode is the Normal Mode, during which the vehicle points the solar panels towards the Sun in a relaxed ±15 deg deadband. All orbit raising / lowering maneuvers are executed using the Delta-V sequencer, which takes the spacecraft from sun pointing to burn pointing through Burn execution and back to Sun pointing. Autonomous transition paths are provided in Normal Mode and the Delta-V Sequencer to handle loss of Sun and / or high vehicle rotation rates. The Descent mode is initiated through a ground command as a part of the pre-descent sequence.

\section{Orbital GNC}
Attitude determination, attitude control and burn execution are the primary functions of the GNC system during the orbital segment of the mission. The spacecraft spends a majority of the time in an inertial pointing mode (Sun pointing). Attitude knowledge is provided by propagation of vehicle rates measured by the IMU, fused with updates from the SSU (updating at 10 Hz) through a Kalman filter. 

\begin{figure}[htbp]
  \begin{minipage}[b]{0.5\linewidth}
    \centering
    \includegraphics[trim= 60mm 20mm 25mm 20mm,clip,width=1.2\linewidth]{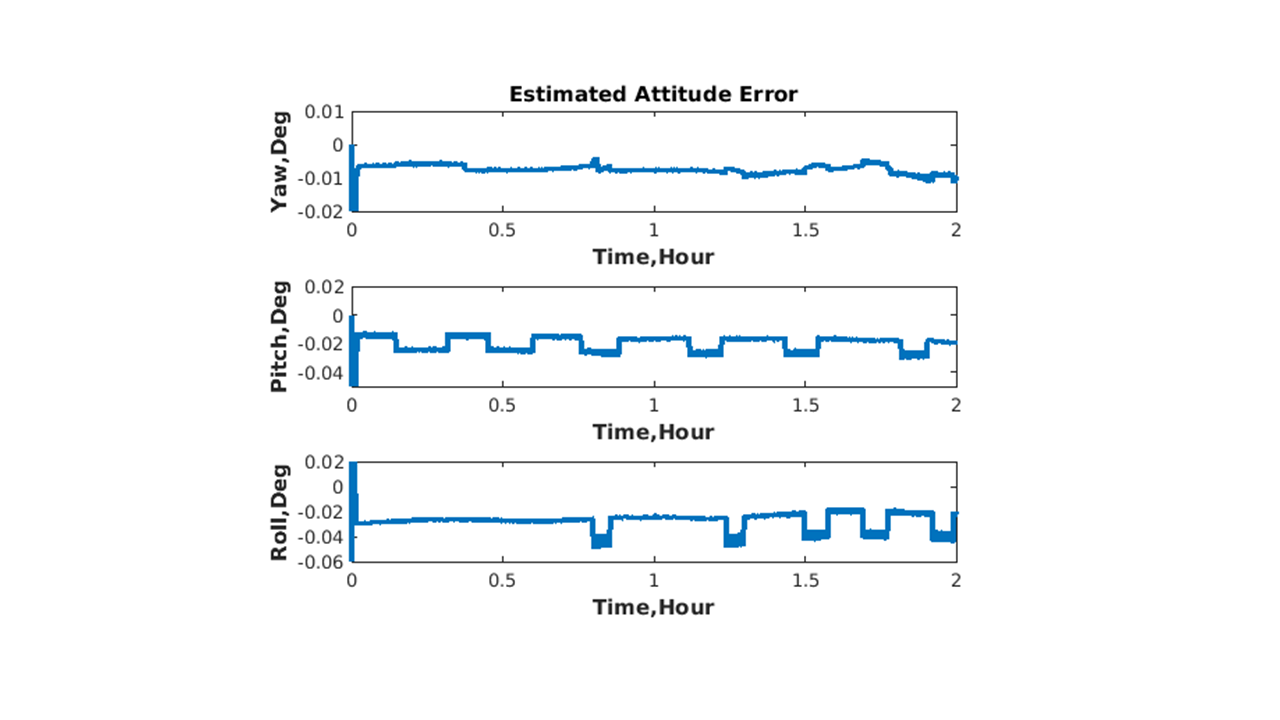}
    \caption{Attitude estimation error}
    
  \end{minipage}
\hspace{0.1cm}
  \begin{minipage}[b]{0.5\linewidth}
    \centering
    \vspace{-10mm}
    \includegraphics[trim= 60mm 20mm 25mm 20mm,clip,width=1.2\linewidth]{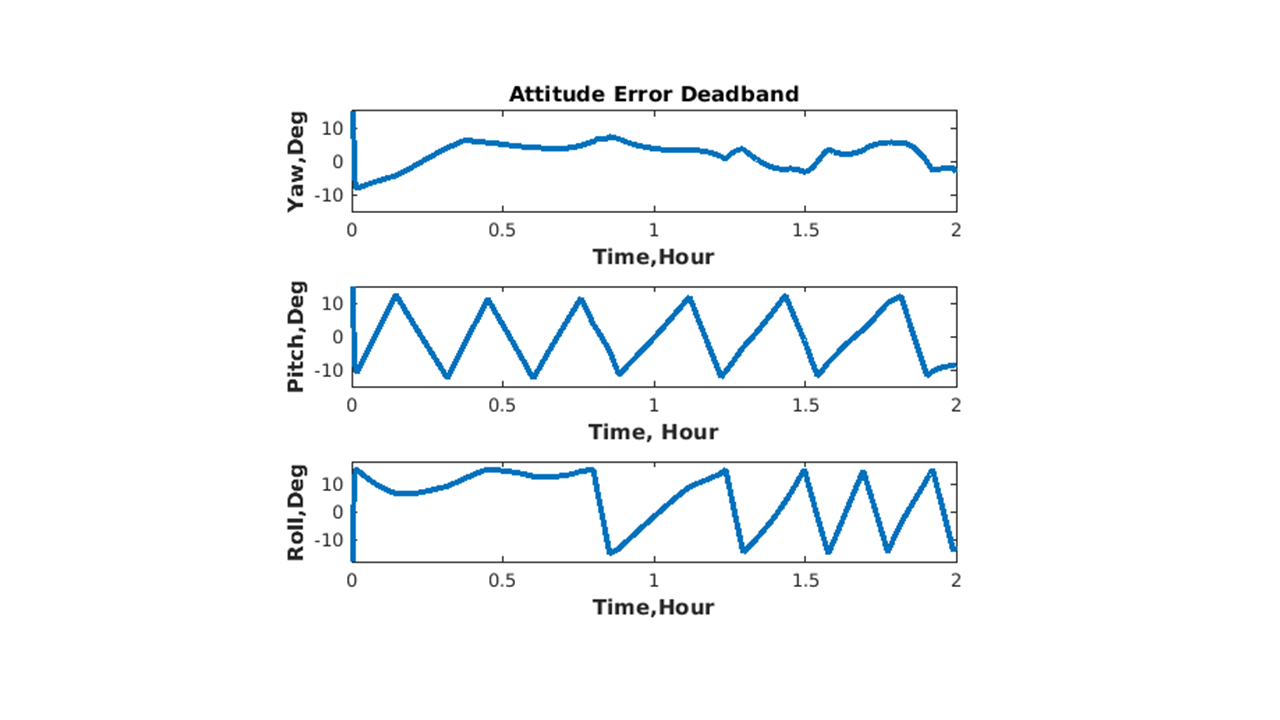}
    \vspace{-8mm}
    \caption{Attitude control error}
   
  \end{minipage}
\end{figure}

Attitude control is achieved using a Proportional-Derivative (PD) controller along with a PWPFM. Thruster firings for attitude control introduce perturbations to the orbit, and hence are desirable to be minimized. Hence, the attitude control is designed to maintain a relaxed ± 15 deg deadband during Sun pointing. In addition, the pulse width of the thrusters are reduced to minimize firings. Figure 8 shows the attitude error profile for a 2-hr period inertial pointing reference. The attitude error is observed to stay within 15 deg. The rapid changes to attitude errors are due to thruster firings at the edges of the limit cycle.

\subsection{DeltaV sequencer}
\begin{wrapfigure}{l}{0.54\textwidth}
  \centering
  \vspace{-9mm}
  \hspace{-10mm}
  \includegraphics[trim=50mm 10mm 45mm 20mm,clip,width=0.6\textwidth]{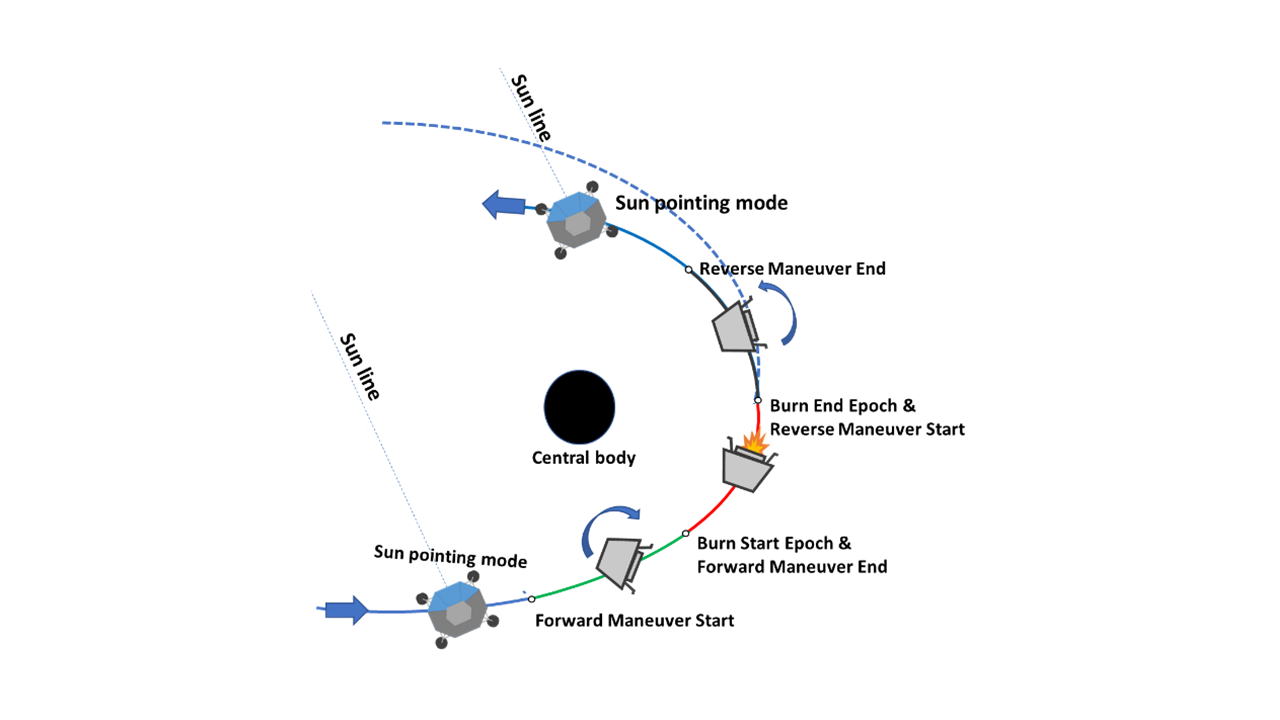}
   \vspace{-10mm}
  \caption{Delta V Sequencer.}
  \vspace{-15mm}
\end{wrapfigure}

The Delta-V sequencer (See Figure 9) is utilized for executing all orbit raising/lowering and Trajectory Correction Maneuvers. The Delta-V sequencer provides a configurable timeline and takes the vehicle from Sun pointing through burn execution and back to Sun pointing without ground intervention. 
Burns can be executed with the main engine or the RCTs alone, and reference attitude  can be configured to be provided by a profile or inertial pointing.

\subsection{Contingency modes}
Contingency modes are used to protect the spacecraft in case of deviations from nominal operation conditions. The Safe Mode is triggered if the Sun vector as derived from the Sun Sensors deviate more than a threshold. On Safe Mode entry, the vehicle locates the Sun in order to maintain a power positive state. The Rate Safe Mode is triggered by persistence of high vehicle rates. All thruster activities are terminated in this mode.

\section{Descent GNC}

The pre-descent sequence configures the spacecraft for lunar descent, with the most important action in this sequence being the initiation of Inertial Navigation. Orbit determination done on the ground provides the position and velocity of the spacecraft at an epoch – chosen to be 10 minutes before perilune. From this epoch onwards, the vehicle position and velocity are propagated using in the Moon Centred Inertial Frame using IMU measurements and an onboard gravity model. 

\begin{figure}
  \centering
  \includegraphics[trim=15mm 5mm 65mm 5mm,clip,width=0.8\textwidth]{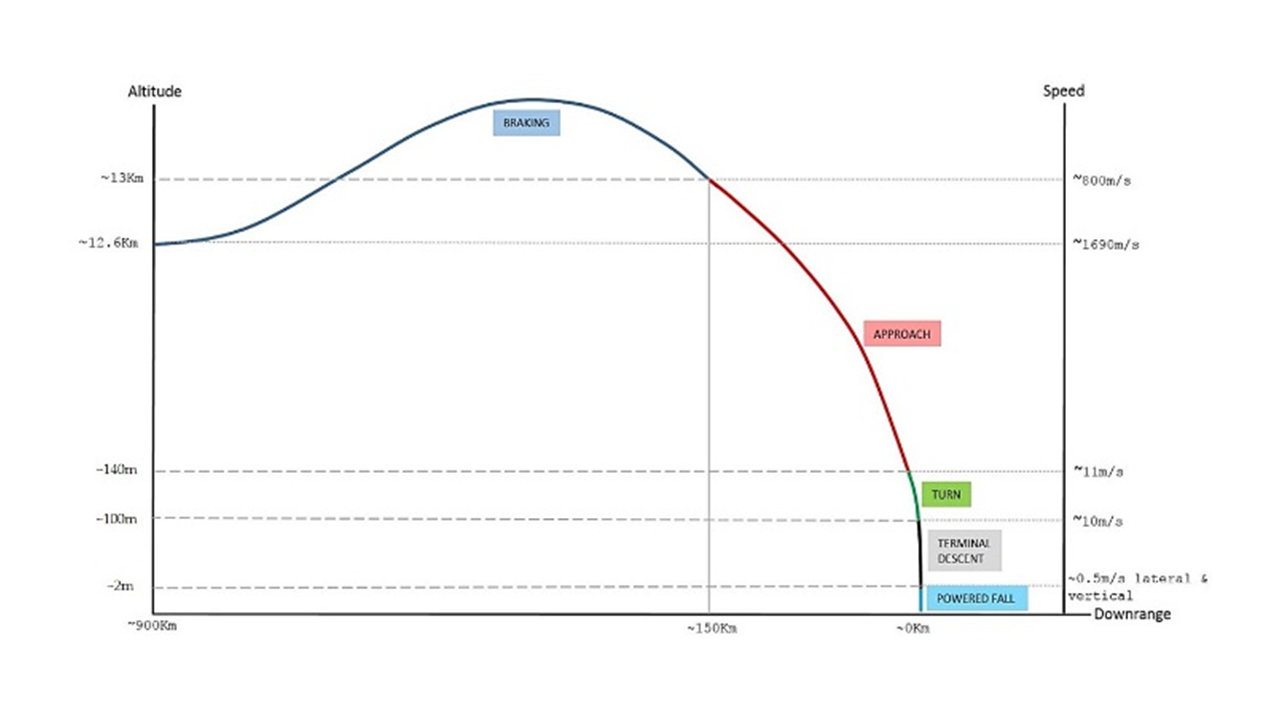}

  \caption{Descent Maneuver Profile.}

\end{figure}

It is to be noted that the target coordinates for the landing are chosen out of a very rigorous process which identifies the point on a Digital Terrain Model which minimizes hazardous areas like slopes, shadows or large craters on its landing site within an expected area of dispersion ellipse for the lander.
The pre-descent sequence also disables SSU updates and orients the spacecraft to the burn start attitude. The entry conditions to autonomous braking locate the vehicle approximately 800 km downrange to the landing site at an altitude of approx. 12.6 km above the datum radius with a speed of 1.7 km/s.  

\subsection{Braking, Approach and Turn phases}
The Braking phase, occurring over 500 seconds and covering 650 km of downrange, is responsible for removing the entry velocity and delivering the vehicle to proper entry conditions for the Approach phase. Navigation information continues to be provided by IMU based propagation, augmented by altitude information using range measurements by the LLRF (at 1 Hz) and an onboard terrain model (See Figure 11). During the Braking phase, the vehicle tracks attitude and thrust profiles uplinked as a part of the pre-descent sequence. These profiles are fuel-optimal and computed on the ground using a pseudo-spectral method based optimal control solver. 

\begin{figure}
  \centering

  \includegraphics[trim = 50mm 20mm 80mm 38mm,clip,width=0.6\textwidth]{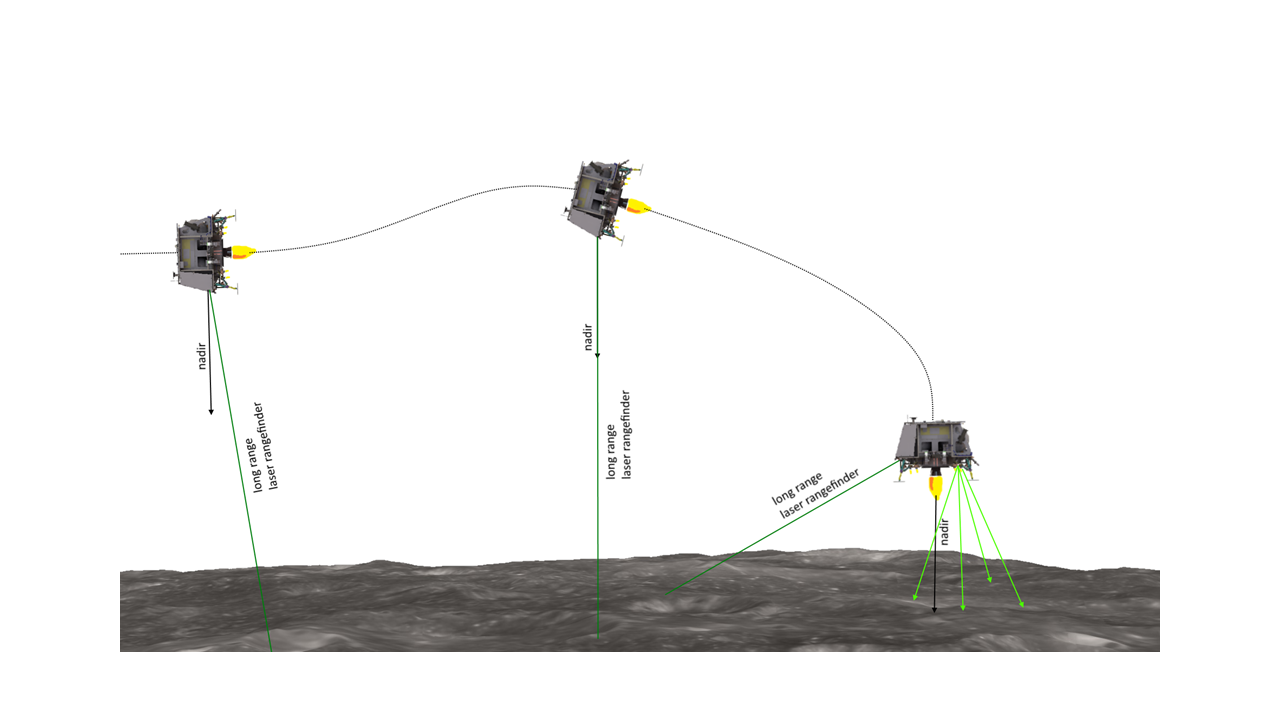}

  \caption{Braking, Approach and Turn phase.}

\end{figure}

Estimates of the vehicle mass and CG are computed using IMU measurements and known propulsion system performance parameters. The Approach phase begins when the vehicle is less than 150 km downrange. The Approach phase of the descent maneuver occurs over 450 seconds and covers the remaining 150 km downrange to the landing site. This phase employs a closed-loop guidance algorithm proposed by D’Souza \cite{Dsouza1997Optimal} to remove all inherited position and velocity errors and brings the spacecraft to desired target conditions for the Turn phase. State information continues to be provided by inertial navigation as in the Braking phase. The Guidance algorithm computes, in real-time, desired acceleration commands to null position and velocity errors relative to target conditions. These accelerations commands are translated to equivalent thrust commands using the vehicle mass estimate provided by the mass estimation logic. During the Braking and Approach phases, the main engine operates in a steady state ON mode. The attitude control thrusters are operated in OFF modulation to provide additional braking thrust and reduce mass consumption. Residual thrust requirements are met by rest of the RCTs operating in pulsed mode. The LDS cameras will be used through the braking phase for capturing images at low frame rates.

The transition to the Turn phase is dictated by speed or time-to-go thresholds. The vehicle enters the Turn phase at an altitude of approx. 150m with near zero terrain relative speeds. Navigation in this phase switches to a sub-satellite East-North-Up frame defined once, and the vehicle executes a pitch maneuver orienting the longitudinal axis vertical. The main engine is turned OFF during this phase to avoid unintentional altitude build-up, and the spacecraft is solely supported by the RCTs.

\subsection{Terminal Descent phase}
\begin{figure}
  \centering
  
  \includegraphics[trim = 20mm 15mm 35mm 20mm,clip,width=0.6\textwidth]{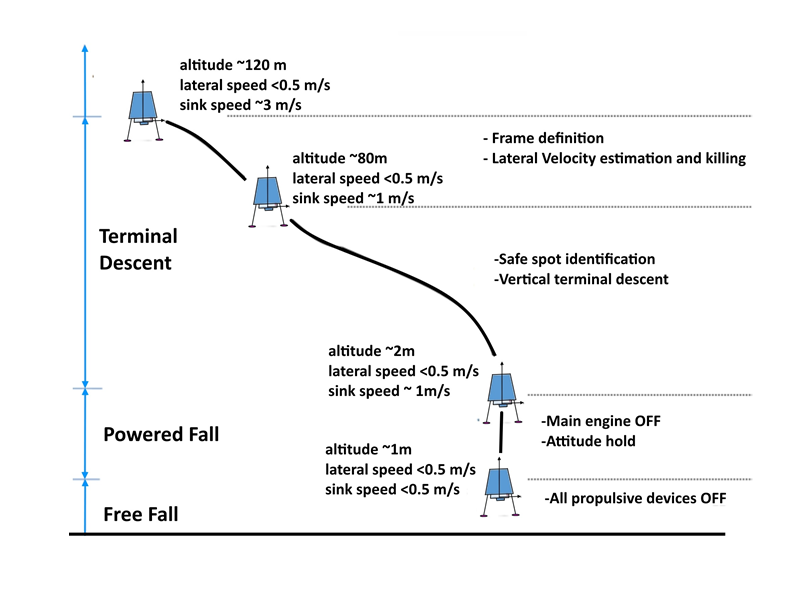}
  \caption{ Terminal Descent Maneuver Profile.}
\end{figure}

The Terminal Descent Phase of descent begins after the turn maneuver and starts approximately at 120m. This phase is the most computationally intense of the entire mission and serves two primary objectives – identification of a safe landing spot and landing with near zero speeds. Figure 12 depicts the maneuver profile of Terminal Descent. After the end of Turn phase, the vehicle follows a constant attitude and sink velocity profile till the short-range LRFs start providing valid range measurements. State information until this point is provided by IMU propagation in the East-North-Up (ENU) frame.

After SLRF acquisition, a local Terrain Relative Navigation (TRN) frame is defined using SLRF measurements (See Figure 13) and used for all subsequent state propagation and image processing tasks. The vertical states - altitude and sink velocity - are estimated using a Kalman Filter. Guidance and Control functions, however, execute in the ENU frame.

\begin{wrapfigure}{l}{0.5\textwidth}
  \centering
  \vspace{-8mm}
  \hspace{-10mm}
\includegraphics[trim=105mm 40mm 0mm 30mm, clip,width=1\textwidth]{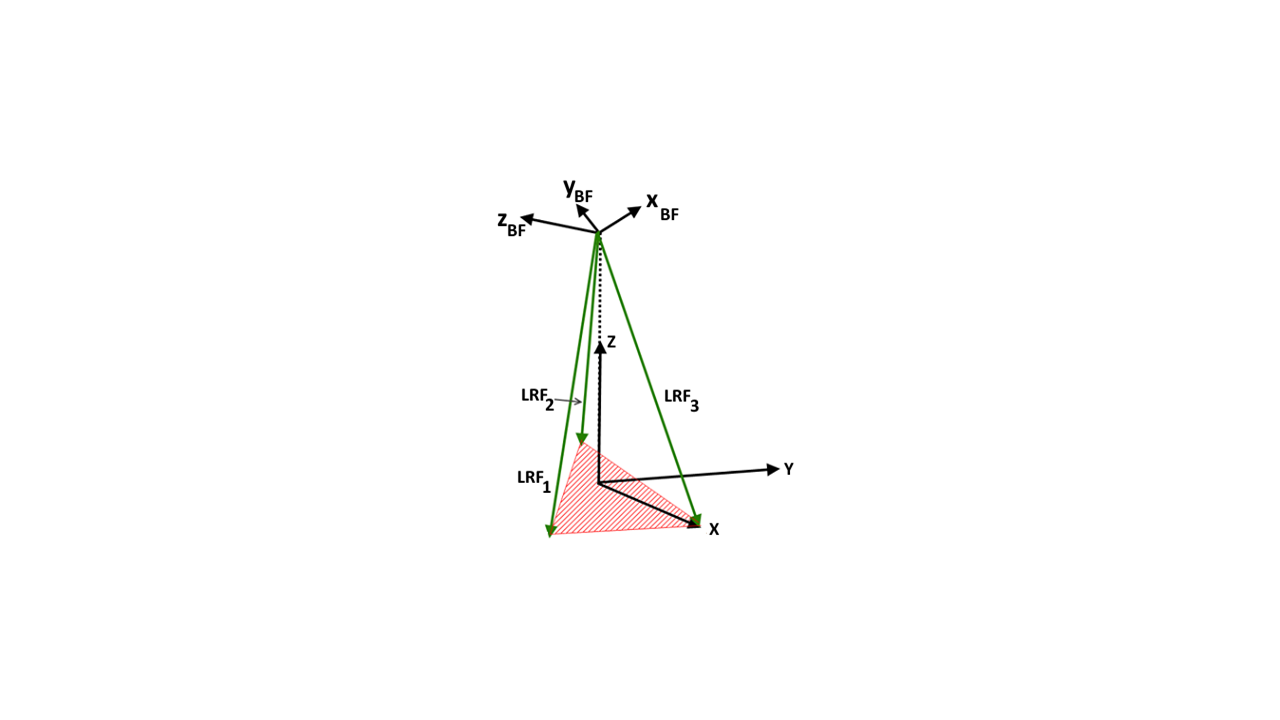}
  \vspace{-8mm}
  \caption{ TRN Frame Definition.}
  \vspace{-5mm}
\end{wrapfigure}

The terrain relative velocity of the vehicle at the start of Terminal Descent inherits errors from the uncertainty of Orbit Determination solution, sensor biases, and misalignment, onboard gravity model uncertainty and propagation errors. In order to ensure soft-landing, the terrain relative lateral velocity of the vehicle is computed by the scale and rotation invariant Speeded-Up Robust Features (SURF) algorithm applied to matched feature pairs from LDS images. The estimated lateral velocity is then removed before initiating Safe Spot Identification (SSID). Safe Spot identification uses images acquired from the LDS to determine, in real-time, a safe landing spot (up to a 20m divert) free of hazards like boulders and craters. Morphological image gradients and local variance-based techniques are used in a spiral search strategy to identify the closest available hazard free region. The decision to divert to the safe spot, however, is conditional on propellant availability. The position and velocity of the vehicle relative to the identified safe spot is continuously provided by combining SURF with a tracking algorithm.  Following a vertical descent segment from 10m, the Main Engine is cut-off at approx. 2m with all RCTs being ON. The latter are cut-off at approx. 1m altitude, and the vehicle freefalls to the surface. During the Terminal Descent phase, the main engine is operated in pulse-mode, with RCTs acting in complement to provide sink velocity control. The vehicle is also rotated along the Z axis to ensure desired landing azimuth to ensure power generation post landing.

It is to be noted that image algorithms like SURF require lighting conditions for the shadows to be projected from objects on the lunar surface. This constraint, along with the need to maximize surface operation duration and power generation dictates the choice of an early morning landing time. 

\section{Operations}
\subsection{Sensor Operation plan}

The operations plan for GNC sensors is depicted in Figure 14. Blank lines indicate that the device is switched off, dotted lines indicate the device is powered on but not being used in the GNC, and the solid lines indicate a sensor powered on and in-the-loop. The IMU, being the fundamental sensor for navigation, will be powered on throughout the mission.

\begin{figure}[h]
  \centering

   \includegraphics[trim=20mm 10mm 20mm 15mm, clip,width=0.6\textwidth]{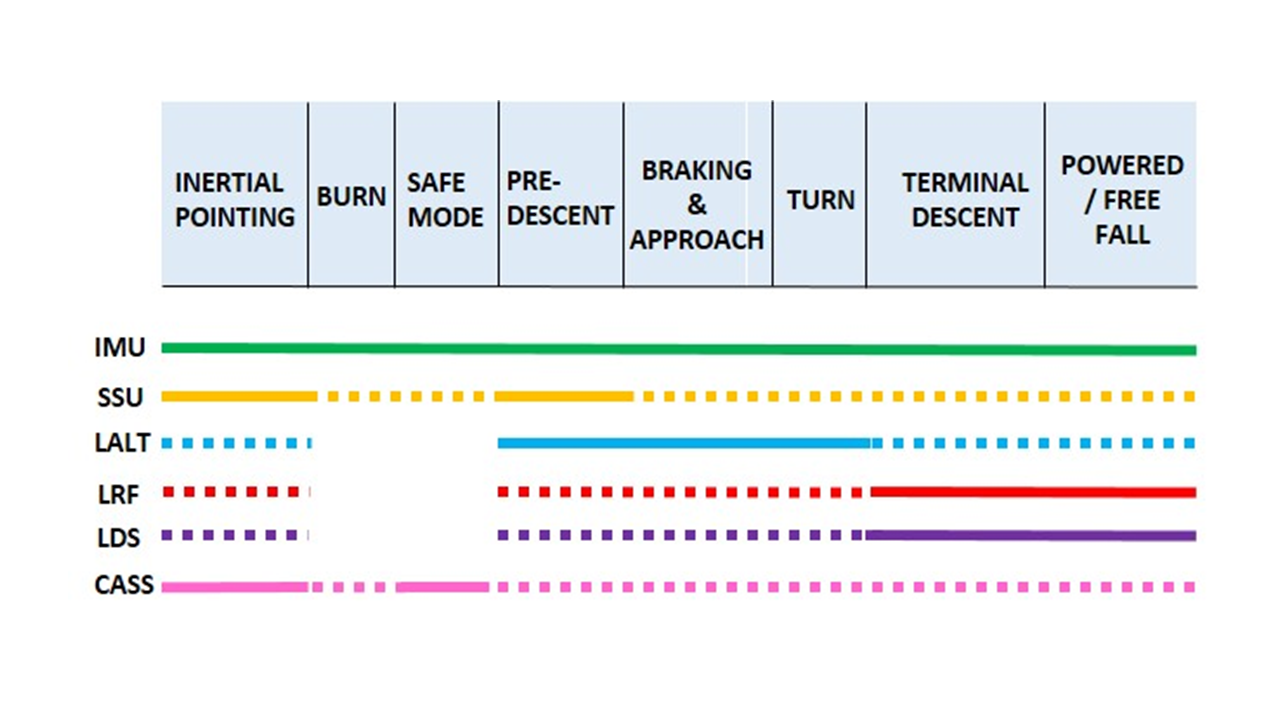}
  \caption{Sensor Operation Plan.}

\end{figure}

The SSU, being a low power consuming device is also powered on throughout the mission, but updates to the navigation loop will be disabled during burns or maneuvers where high attitude rates are expected to avoid high measurement errors. A consistency logic checks for mutual agreement between IMU derived attitude and that provided by the SSU. Sun sensors, being passive devices, are always available to provide Sun pointing reference attitude. 

\begin{figure}
  \centering

  \includegraphics[trim=20mm 10mm 40mm 15mm, clip,width=0.6\textwidth]{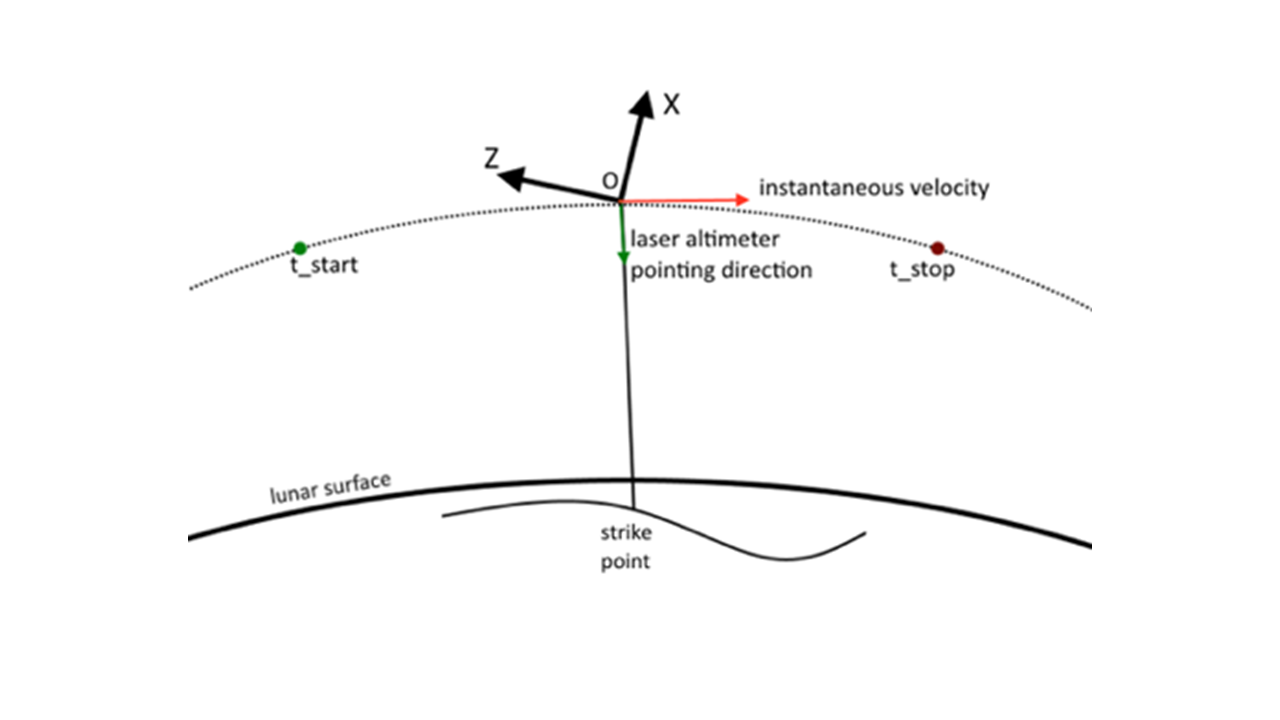}
  \caption{PSPO LLRF checkout.}

\end{figure}

\subsection{Orbit navigation}
  During the cruise phase of the lunar transfer trajectory, for flight path control, three TCMs will be allocated to achieve nominal arrival conditions at the moon. The first TCM is executed 24 hours after TLI to correct the dispersions due to TLI burn performance. The second and third TCMs will be executed if there are specific delta-V correction thresholds crossed. The orbit determination for an orbit of size 12.6 x 100 km around the Moon is presented. The complete simulation span is 5 hours and the orbit complete 2.5 revolutions within this span. The tracking schedule comprises of 10 minutes of ranging and 1 hour of two-way doppler data.  The sampling interval is 10 seconds. The residuals are checked if they are within ± 3 sigma values and the filter-smoother consistency tests to validate the filter estimates. The filter-smoother state differences are divided by the filter-smoother variance differences. For a sample mission, the difference in position and velocity between the truth and the estimated trajectory is 7 m and 9 mm/s. The position uncertainty (1 sigma) in RIC coordinates is Radial:7.5 m Intrack:160 m and Cross-track:35 m and the velocity uncertainty (1 sigma) in RIC coordinates is Radial:15 cm/s Intrack:0.6 cm/s and Cross-track:7.8 cm/s. 

\section{Simulation}
\subsection{Overview}

Simulations are utilized extensively to demonstrate the effectiveness of the GNC system to meet mission requirements. All GNC simulations are carried out using an in-house developed simulation framework called  TeamIndus Guidance navigation and control EmulatoR (TIGER) developed in the MATLAB/Simulink environment.

This framework is used for trade studies, non-real time simulations, real-time Processor-In-Loop simulations as well as Monte-Carlo runs. The simulation environment includes models of vehicle physics, slosh, sensors, and actuators as well as environment models. The following sections describe these models along with typical simulation results.

\subsection{Modeling}
The modeling section covers critical physical models used or developed for the GNC simulations. 
\subsubsection{Gravity}
The gravity model selected is the GRAIL 100x100 gravity model. The gravity modeling error is kept at 95 mgal, which is the maximum error between the LPK100K, SGM90d and SGM100h models \cite{terster1997nasa}. 

\subsubsection{Slosh}

\begin{wrapfigure}{r}{0.6\textwidth}
  \centering
  \vspace{-5mm}
  \hspace{-30mm}
  \includegraphics[trim=70mm 25mm 15mm 15mm,clip,width=0.8\textwidth]{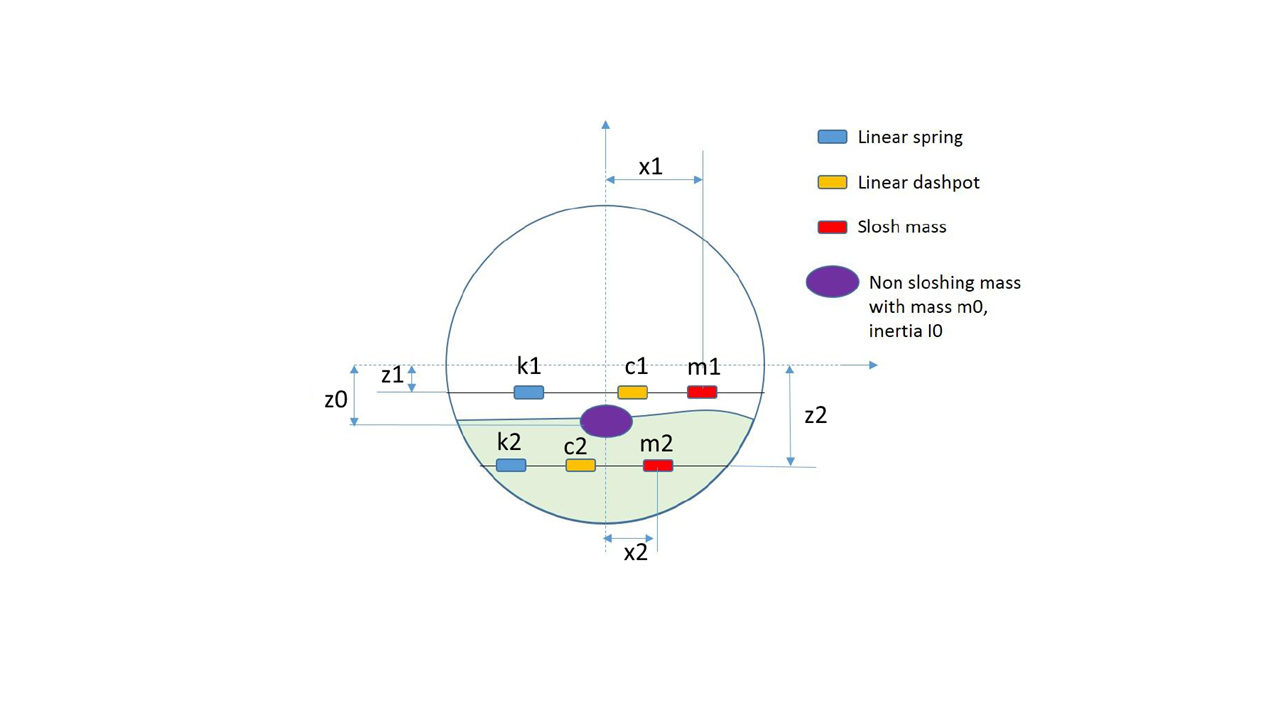}
  \caption{Slosh Modeling in a Tank}
  \vspace{-5mm}
\end{wrapfigure}

Liquid sloshing is a phenomenon that occurs when the liquid in a partially filled tank moves around and exerts disturbance force and moments on the spacecraft and significant during periods of the mission that have significant thrusting activity (Delta-V maneuvers and Descent). Although the liquid free surface oscillation has various modes, the first lateral mode is responsible for a significant amount of disturbance forces. An equivalent mechanical model consisting of linear springs, dampers, slosh masses and fixed masses is used to model these dynamics, as shown in Figure 16. Since the effects of lateral slosh are strongly influenced by accelerations imposed on the vehicle and the fill fractions, the parameters of such equivalent mechanical models vary as a function of imposed acceleration and fill fraction.

\subsubsection{Actuator}
The true thrust models for each actuator is built separately to reflect the command-to-actuation latency, thrust rise and decay characteristics for each thruster. The engines used in the TeamIndus mission have undergone extensive acceptance tests to provide thrust characteristics at various modes of operation. The basic strategy is to model thrust as a combination of three second-order transfer functions and a first-order lag and then tune the transfer functions to get a best fit given the data. Also, thrust and Isp have been modeled as a function of the thruster duty cycle separately. 

\subsubsection{Landing site terrain modeling}
As discussed previously, the TeamIndus lander uses image processing algorithms during the Terminal Descent phase of lunar descent to estimate lateral velocity and identify a safe landing spot. In-the-loop evaluation of these algorithms necessitate generation of appropriate scenes of the environment. The scene thus generated should emulate: a)	The observed terrain geometry and corresponding statistics which include terrain slope/roughness characteristics, crater Size-Frequency Distribution (SFD) and boulder SFD b)	The observed terrain optical properties which are lunar regolith albedo, irradiance intensity, incident light spectral characteristics and lunar surface reflectance.

\begin{figure}
  \centering
  \includegraphics[width=0.7\textwidth]{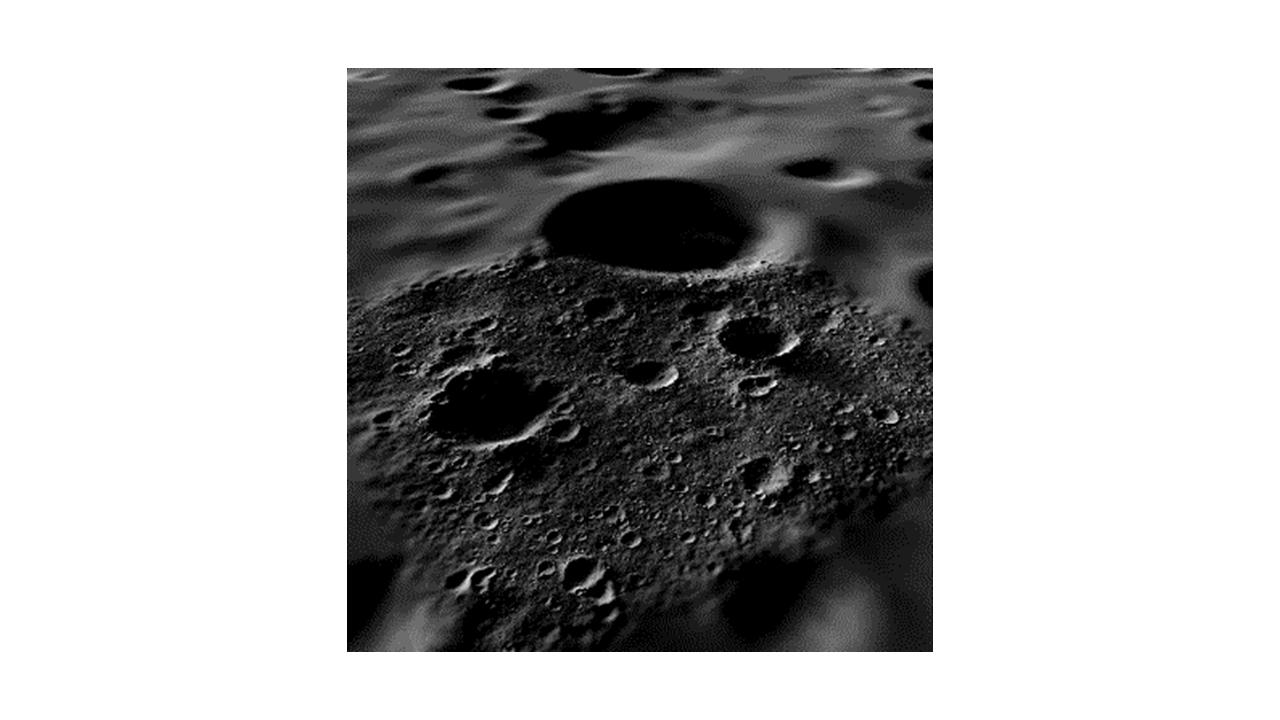}
  \caption{Super-resolved 0.02m x 0.02m resolution terrain on PDS imported 5m x 5m resolution}
\end{figure}

To this end, the Planet and Asteroid Natural Scene Generation Utility (PANGU) \cite{Parkes} is utilized to provide representative images and range measurements of the SRLFs. The DTM containing the landing site \texttt{NAC\_DTM\_IMBRIUM\_E287N3336}  from PDS Geosciences \cite{Henriksen}, has a resolution of 5 meters and forms the base mesh layer.  The super-resolution algorithm is employed to enhance resolution from 5 meter per pixel to 0.05 meter per pixel. In our simulations, the tabulated data sizes for crater diameters from 3.9m to 8000m (as bins spaced logarithmically) have been taken from Hartman Production Function (HPF) which has been recorded \cite{luna1969Hut}. The boulder distribution is also represented as power SFD. In this case, we have chosen the boulder SFD from \cite{slyuta2015dis} as the carpet distribution for boulder above 1.5m diameter. Figure 17 shows a sample scene generated from PANGU. The distance of sun along with the irradiance values are used to adjust photon counts to tweak the surface appearance to look as close as possible to existing lunar imagery. Further, the surface reflectance is modeled by Hapke parameters, values of which for lunar mare have been obtained from existing literature.

\subsection{Delta V sequencer simulation}

Figure 18 shows the attitude error profile for delta V burn maneuver. The attitude error is observed to stay within +/-15 deg during Sun pointing, +/-1 deg during the forward maneuver and reverse maneuver and within +/-- 0.2 deg during the burn.

\begin{figure}[htbp]

  \begin{minipage}[b]{0.5\linewidth}
  \vspace{-11mm}
    \centering
     \includegraphics[trim=60mm 20mm 70mm 20mm,clip,width=1\textwidth]{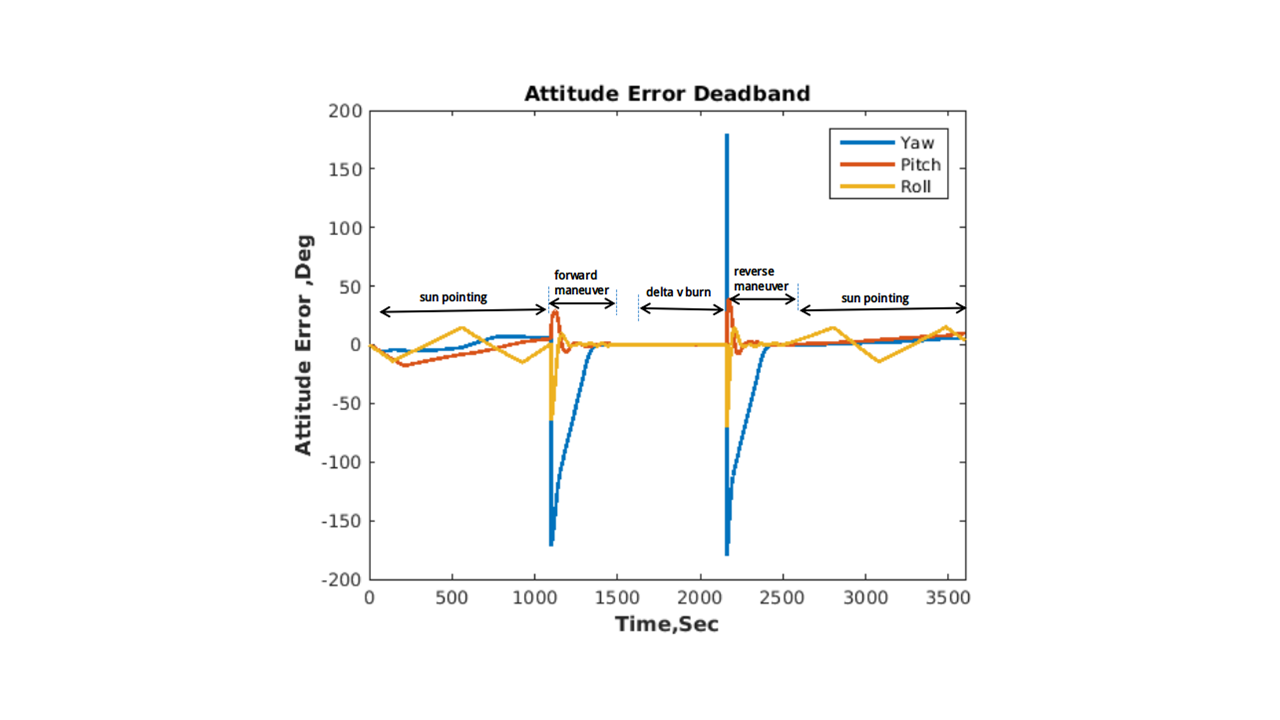}
      \caption{Attitude error in a Delta V Sequence}
    
  \end{minipage}
  \hspace{0.2cm}
  \begin{minipage}[b]{0.5\linewidth}
    \centering
    \includegraphics[trim=60mm 10mm 50mm 15mm,clip,width=1\textwidth]{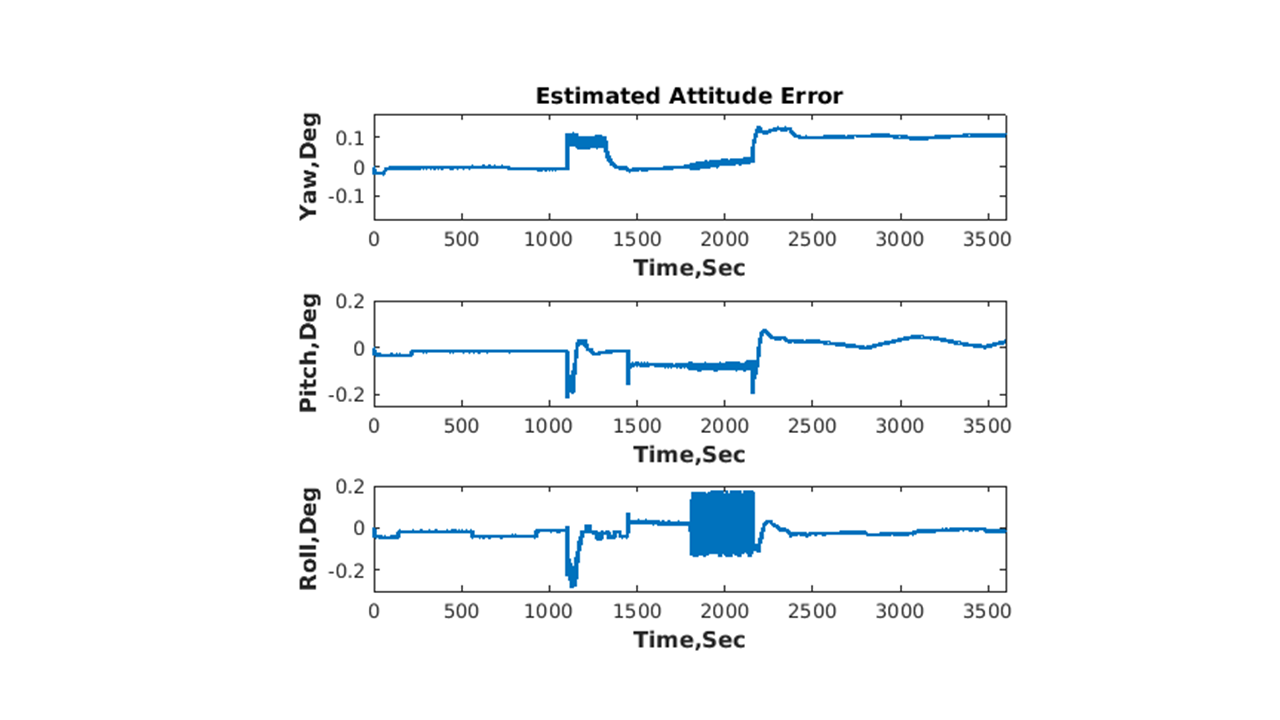}
    \caption{Attitude estimation error in Delta-V sequence}
   
  \end{minipage}
  \vspace{-5mm}
\end{figure}

The burn is terminated based on the accumulated delta-V reported by the accelerometer or a timer, with the former given preference.Figure 19 shows the attitude estimation error during delta-v sequencer operation. During the burn, star sensor updates are off, hence estimation error observed is +/-0.2 deg.

\subsection{Braking, Approach and Turn phase simulation}

\begin{figure}[h!]
  
  \centering
  \includegraphics[trim=20mm 40mm 0mm 25mm,clip,width=1.0\textwidth]{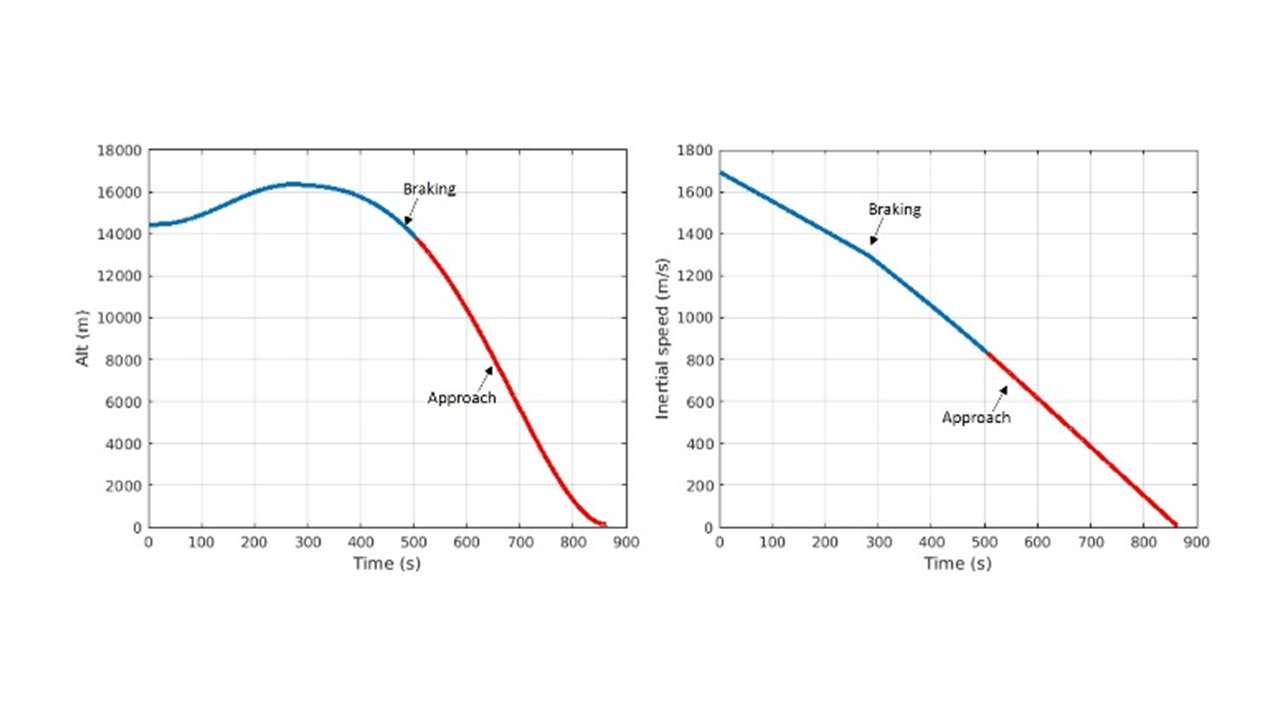}
  \caption{Altitude and Speed during Braking and Approach}
  \vspace{-5mm}
\end{figure}

The left plot in Figure 20 shows the time history of the vehicle altitude during the Braking and Approach phases. The spacecraft starts at approx. 15km (relative to landing site elevation) and, over a period of approx. 850s, reduces the altitude to approx. 150m. The right plot shows the inertial speed of the spacecraft. It is observed that the entry orbital speed is reduced by approx. half during the Braking phase, and the rest is removed in the Approach phase.

\subsection{Terminal Descent Phase simulation}
\begin{figure}[htbp]

  \begin{minipage}[b]{0.5\linewidth}
    \centering
    \includegraphics[trim=40mm 2mm 40mm 5mm,clip,width=1.1\linewidth]{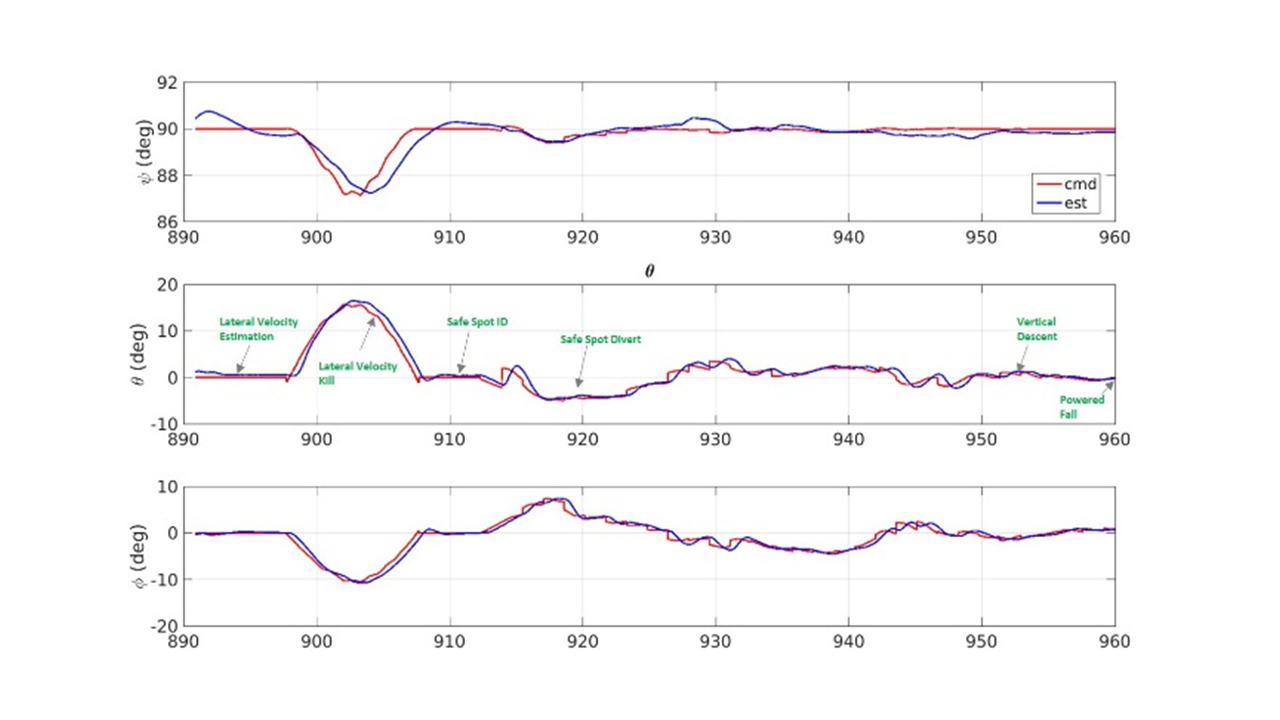}
     \caption{Terminal Descent Attitude Evolution }
    
  \end{minipage}
  \hspace{0.5cm}
  \begin{minipage}[b]{0.5\linewidth}
    \centering
    \includegraphics[trim=40mm 10mm 40mm 5mm,clip,width=1.1\linewidth]{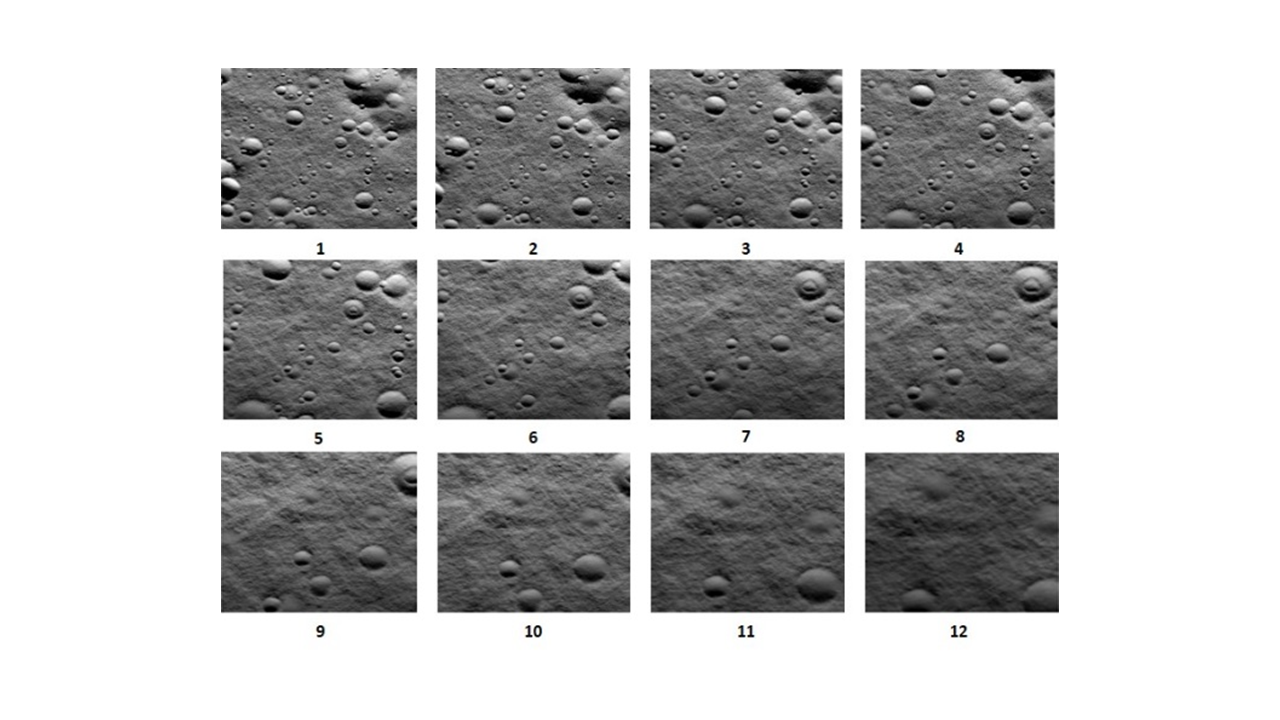}
    \caption{Shows a sequence of images from a Terminal Descent simulation. }

  \end{minipage}
  \vspace{-5mm}
\end{figure}

   The red line in Figure 21 indicates the commanded attitude while the blue line indicates the attitude estimate. The various segments during this phase are also highlighted. The significant tilt of the vehicle longitudinal axis from the local vertical is observed for the lateral velocity kill and the safe spot divert maneuvers.

\begin{figure}[h!]
  \centering

  \includegraphics[trim=0mm 30mm 10mm 25mm,clip,width=1.0\textwidth]{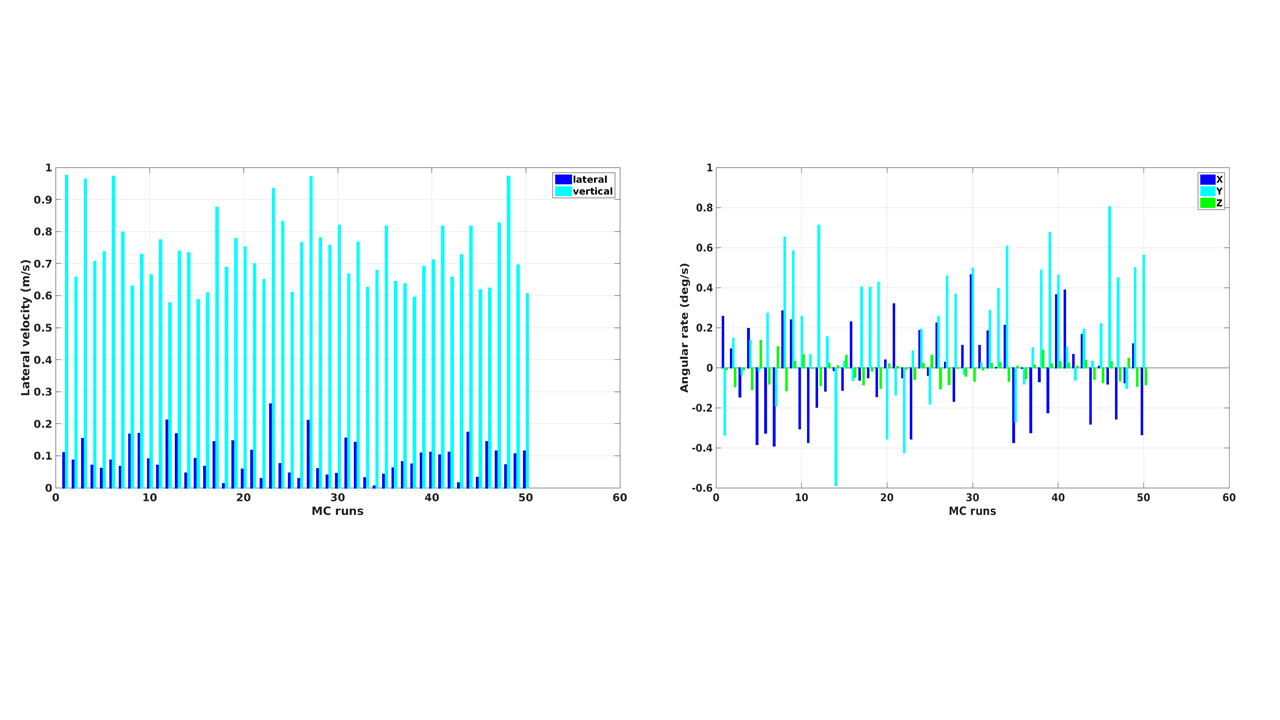}
    \vspace{-13mm}
  \caption{Touchdown conditions}
  \vspace{-5mm}
\end{figure}
 Figure 22 shows a sequence of images as viewed by one of the LDS cameras from a terminal descent simulation. As of the first image, the vehicle has identified a safe spot and executed a divert maneuver. Image numbers 1 – 12 show the vertical descent segment that brings the vehicle to the identified safe landing spot.\\
Figure 23 shows the spread of touchdown speeds and body rates for a typical set of Terminal Descent Monte-Carlo runs. It is seen that the body rates are well within 1 deg/s at touchdown, while the 0.5 m/s and 1 m/s are the upper bounds for the lateral and vertical touchdown speeds.

\subsection{Monte Carlo runs for landing dispersion}
\begin{wrapfigure}{l}{0.5\textwidth}
 \vspace{-5mm}
  \centering
   \hspace*{-5mm}
  \includegraphics[trim=30mm 20mm 20mm 15mm,clip,width=0.6\textwidth]{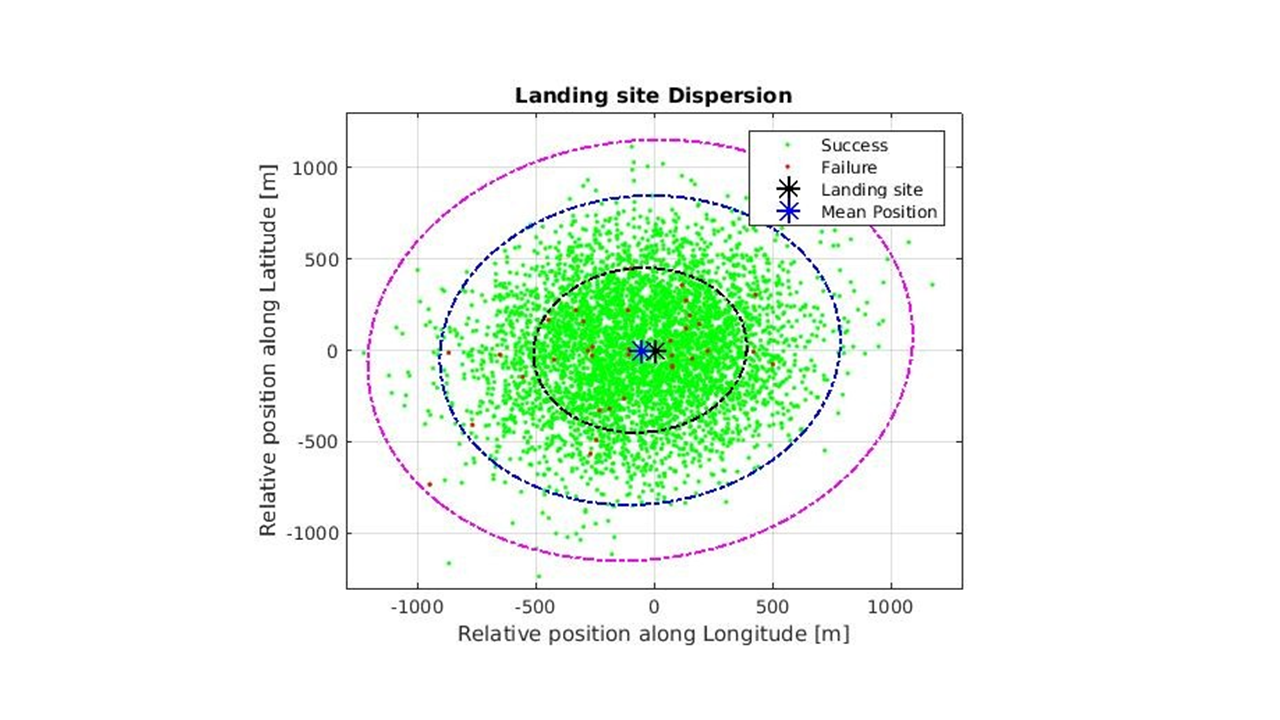}
  \caption{Landing Site Dispersion}
  \vspace{-5mm}
\end{wrapfigure}
The performance of the GNC system during lunar descent maneuver is evaluated using Monte-Carlo simulations. Various parameters that affect the maneuver are varied using their probability distributions, are as follows:

1.	Initial s/c true position and velocity dispersions are modeled as 3 sigma Gaussian distribution errors.
2.	Initial true attitude and attitude rate errors are modeled as a uniform distribution.
3.	The initial mass of the s/c varied in a range based on the min and max possible descent start arrival mass.
4.	The sensor placement/misalignment error for IMU and the laser altimeter is also included in the simulation.
5.	Engine performance errors in terms of the delivered thrust and OFR ratio for the Main Engine and the ACTs are also modeled.

A typical set of criteria for declaring success / failure are:
-	Fuel and Ox and end of Approach > threshold
-	Altitude at end of Approach > threshold
-	Touchdown velocities < threshold
-	Vehicle rotation rates at touchdown < threshold
-	Dispersion w.r.t. desired landing site < threshold
-	
Landing site dispersion for 5000 Monte-Carlo descent simulations are shown in Figure 24.  The three ellipses represent 68.27, 95.45 and 99.73 percentage dispersion boundaries respectively.

\section{Overview}

\subsection{Verification and Validation}
\begin{wrapfigure}{r}{0.7\textwidth}
  \vspace{-8mm}
  \centering
  \hspace{-5mm}
  \includegraphics[trim=25mm 45mm 25mm 25mm,clip,width=0.7\textwidth]{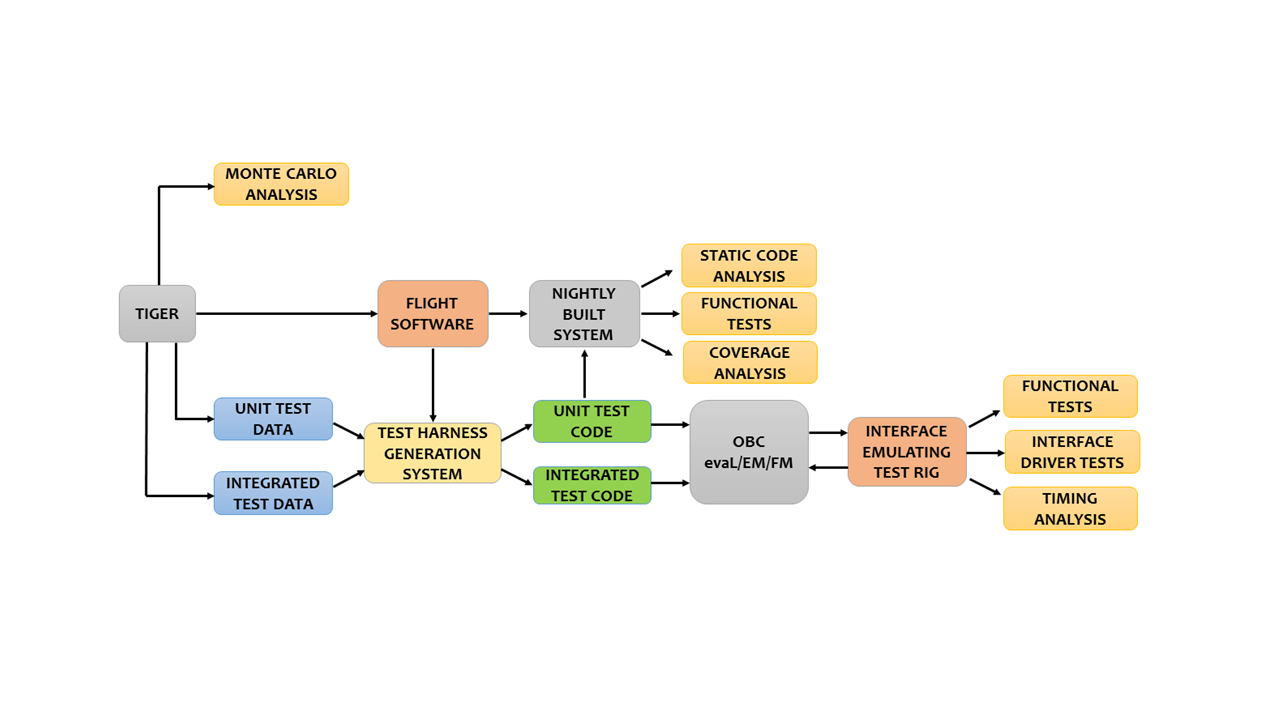}
  \caption{Software Test Process}
  \vspace*{-5mm}
\end{wrapfigure}

Testing and Verification procedures are critical to ensuring that the flight software performs as intended. The verification process for the TeamIndus lander flight software is done at multiple levels, each with increasing fidelity, culminating at mission rehearsals. Figure 25 depicts the software testing process used for validation of flight software functionality. A high level of automation, from test generation through execution and report generation is required to ensure integrity and repeatability of test results. Unit and Integration tests form the source of truth for the flight software, which is executed on hardware and associated flight representative test harnesses. These tests, along with static code and coverage analyses are run as a part of a nightly build system. This process allows teams to be responsive to updates and error fixes.

\subsection{Processor-in-Loop simulations}
The Processor-In-Loop-Simulations (PILS) is the last stage of the testing process and are used for end-to-end mission rehearsals using flight representative hardware, software (onboard and ground) and personnel.
\begin{wrapfigure}{r}{0.7\textwidth}
\vspace*{-5mm}
\hspace{-5mm}
  \centering
    \includegraphics[trim=40mm 10mm 15mm 20mm,clip,width=0.8\textwidth]{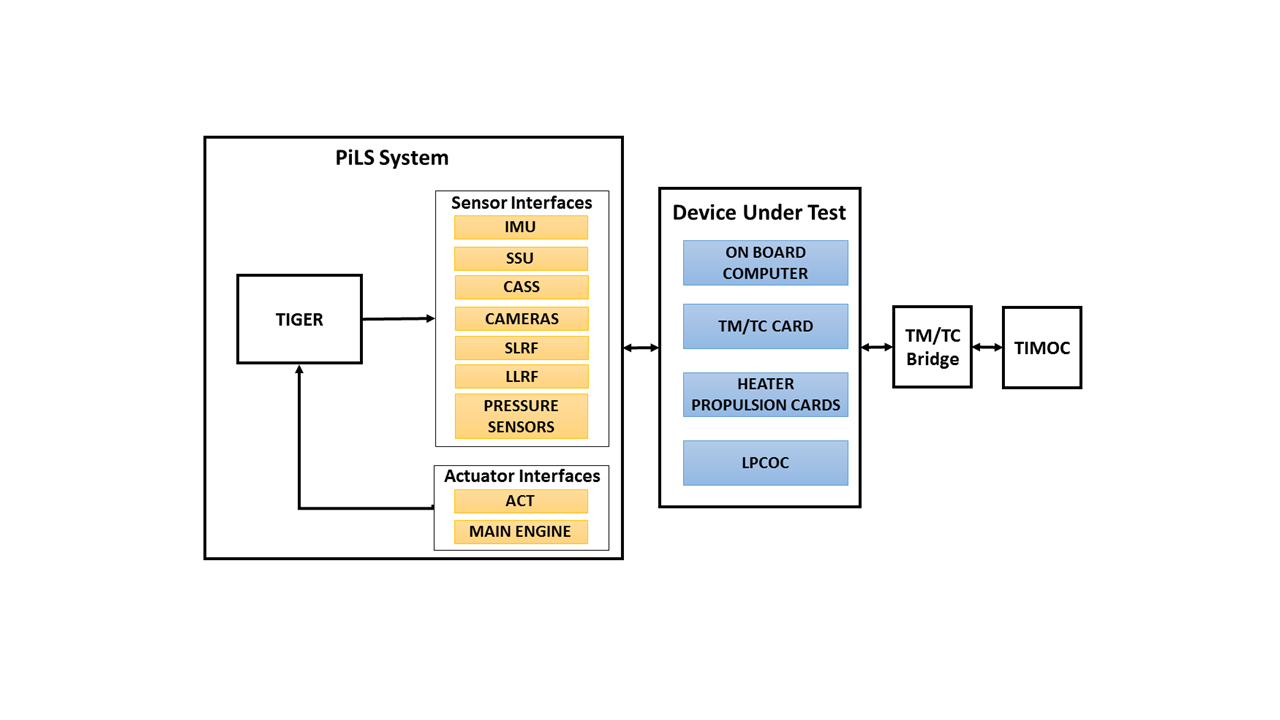}
    \vspace{-18mm}
  \caption{Processor-In-Loop Simulations}
  \vspace{-5mm}
\end{wrapfigure}
The key objectives of such exercises are:
a)	Exercise and validate nominal and contingency flight control procedures.
b)	Validate the interfaces between the spacecraft, ground station and the operators.
c)	Validate data delivery timelines between various teams involved in maneuver planning and execution.

Figure 26 depicts the setup used for PILS / Mission rehearsals. The TIGER framework (See section 9.1), executing in a real-time test system, is used to emulate the vehicle motion, sensors, actuators and the environment. The avionics unit consists of the OBC, the HPCs, TM/TC card and the LPCOC, and receives sensor data from the test system using flight representative interfaces and transmits actuator commands back to the TIGER framework. The TeamIndus Mission Operations Center (TIMOC), containing the mission ground software and the operations team closes the loop. Mission rehearsals of various scenarios are carried out periodically using the PILS system and exercise the end-to-end chain comprising the flight control procedures, mission ground software, spacecraft Tele-Command and Telemetry handling and flight software.  

\section{Risk mitigation studies}
\subsection{Sensor failure scenarios}
	There are a limited set of sensor failures that can be tolerated to support the complete set of mission objectives. Clearly, if the IMU or the Star sensor fail during any point in the mission, it would be a potential mission failure. Failure of the sun sensor system is not critical to the mission and safe mode transitions due to sun sensors can be disabled. Health checks for sensors are performed before the lunar descent. If one of the redundant LLRFs is identified as failed, the active sensor index will be telecommanded so that only the healthy sensors operate in critical phases. For the failures in LDS or SLRFs, the program has inbuilt fault identification and recovery algorithm which mitigates the risk.

\subsection{Actuator failure scenarios}
\begin{wrapfigure}{r}{0.5\textwidth}
  \centering
  \vspace{-9mm}
  \hspace*{-12mm}
  \includegraphics[trim=5mm 5mm 5mm 10mm,clip,width=0.7\textwidth]{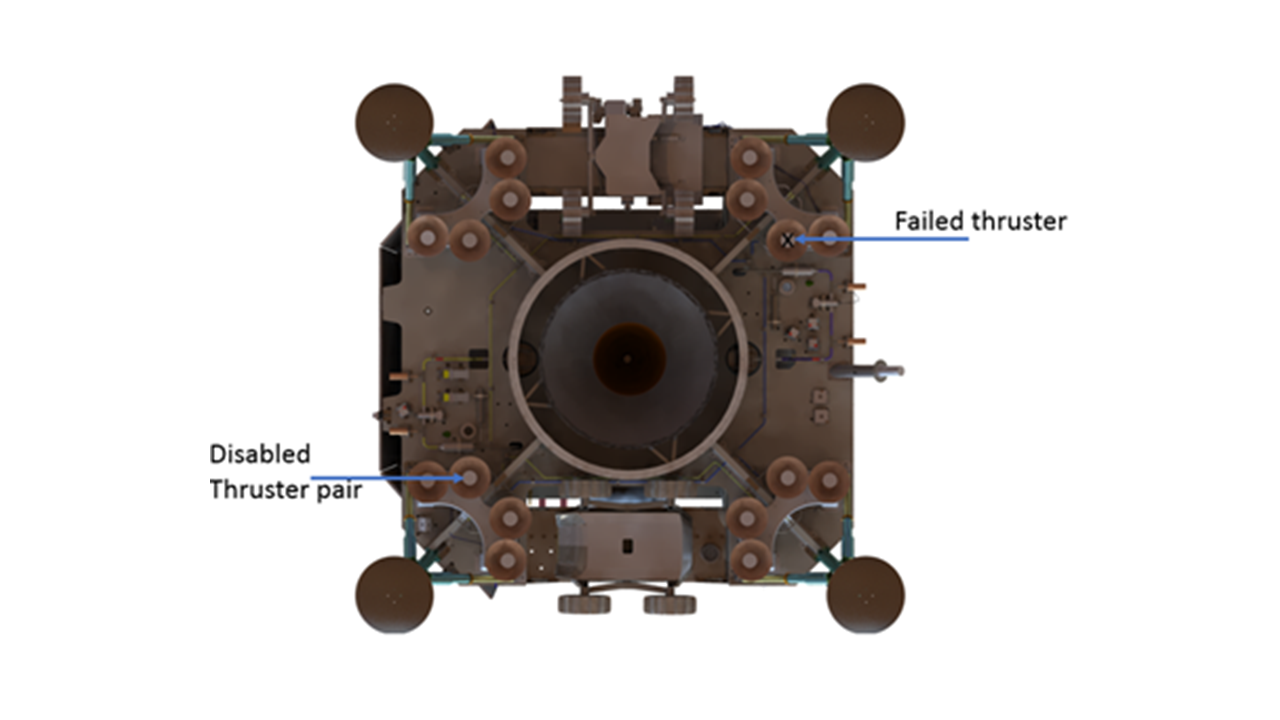}
  \caption{Actuator failure scenario}
\end{wrapfigure}

Due to the short mission duration, the system design accounts for a single RCT failure scenario. If one of the attitude control thrusters fail, the block of four thrusters are disabled and a redundant block is used instead. In the case of a non-canted thruster failure, the matched thruster is also disabled, as shown in Figure 27.	In the interest of algorithmic simplicity and short maneuver duration, no active fault detection and reconfiguration of the actuation system is planned during descent.

\subsection{Main engine cutoff altitude}
A key risk in terminal descent is the operation of the main engine very close to the lunar surface. The plume exhaust may lead to severe regolith erosion and a significant volume of dust particles to fly off and settle on the solar panels, adversely affecting solar power generation post landing. It is observed from data derived from sources [9,10] that the maximum surface pressure exists at the intersection of nozzle axis with the surface for nozzle heights greater than 2 times the nozzle exit diameters (De). For the main engine BT4, the (h/De) of 2 for the test model corresponds to 51cm, below which the nozzle jet would interact with the lunar surface, causing greater erosion and particle dispersion which will be damaging to the engine also and will impact its performance. For the spacecraft main engine, the data suggests ~0.5m is minimum height below which erosion tends to become very high. Taking a very conservative safety margin, engine cut-off height is recommended to be ~2m. 

\section{Summary and conclusion}
The paper attempted to provide an overview of the design of the Guidance, Navigation and Control subsystem and associated aspects of risk reduction. Mode description, attitude estimation, lunar descent phases and sensor operation plans have been described. Orbit navigation, modeling of environment and spacecraft dynamics along with simulation results have been provided. These are some key points which determine mission probability of success. Software testing procedure has been described. Each flight critical hardware will be up-screened for making it failure-proof for the duration of the intended mission. 

TeamIndus has developed significant GNC experience in:

1. Requirements based design philosophy
2. Risk analysis and mitigation
3. Verification and validation experience including Processor in the Loop simulation for GNC software
4. Rigorous landing site selection
5. Cost-effective GNC architecture
6. Unique GNC strategy and simulation framework:
-	Innovative mass estimation scheme
-	Modeling spacecraft dynamics and environment especially for the moon 
-	Innovative scheme for pulsed mode operation of the main engine in the terminal descent. Tested profiles 
   in hot firing tests. 
-	Unique hazard avoidance and lateral velocity estimation scheme for terminal descent 
-	Automated DeltaV sequencer 
-	Autonomous Descent sequencer

\section*{Acknowledgments}
TeamIndus is deeply indebted to NASA Technical Reports Server (NTRS) and Lunar and Planetary Institute for sharing invaluable information on the Apollo and Surveyor missions. We are grateful to Vinayak Vadlamani and Udit Shah for leading the early GNC design, Nitai Agarwal and Nitish Singh for the work done in the field of vision-based navigation, Adithya Kothandhapani for pushing us document system requirements and risk, Jim Fletcher for the Orbit determination solutions, Genoedberg Nadar for generating trajectory solutions in time, Ananth Ramesh for the vibration analyses, Nakul Kukar for the plume impingement study, Rakeshh Mohanarangan for the study of the plume refractive index and Mardava Gubbi for his efforts to automate the test process and bring greater cooperation between GNC and FSW teams. Inputs and support from professor Radhakant Padhi were very valuable from the start. Heartfelt gratitude is due for Professor R.V. Ramanan, Dr. A. Pashilkar, Mr. Srinivas Hegde, and Professor V.S. Adimurthy for reviewing our design and extending insights about real mission constraints.  

\bibliography{sample}

\end{document}